\documentclass{article}

\usepackage{microtype}
\usepackage{graphicx}
\usepackage{subfigure}
\usepackage{booktabs} %

\usepackage[hyphens]{url} %
\usepackage{hyperref}

\usepackage{verbatim}
\usepackage{xspace}
\usepackage{xcolor}
\usepackage{booktabs}
\usepackage{graphicx}
\usepackage{makecell}
\usepackage{amsmath}
\usepackage{enumitem}
\usepackage{nccmath}
\usepackage{mathtools}

\newcommand{\batchreuse}{\textit{RecD}\xspace}
\newcommand{\scribe}{\revised{Scribe}\xspace}
\newcommand{\tectonic}{\revised{Tectonic}\xspace}

\newcommand{\revised}[1]{#1}

\usepackage[accepted]{mlsys2023}

\mlsystitlerunning{\textit{RecD}: Deduplication for End-to-End Deep Learning Recommendation Model Training Infrastructure}

\begin{document}

\twocolumn[
\mlsystitle{\textit{RecD}: Deduplication for End-to-End Deep Learning Recommendation Model Training Infrastructure}

\mlsyssetsymbol{equal}{*}

\begin{mlsysauthorlist}
\mlsysauthor{Mark Zhao}{stanford}
\mlsysauthor{Dhruv Choudhary}{meta}
\mlsysauthor{Devashish Tyagi}{meta}
\mlsysauthor{Ajay Somani}{meta}
\mlsysauthor{Max Kaplan}{meta}
\mlsysauthor{Sung-Han Lin}{meta}
\mlsysauthor{Sarunya Pumma}{meta}
\mlsysauthor{Jongsoo Park}{meta}
\mlsysauthor{Aarti Basant}{meta}
\mlsysauthor{Niket Agarwal}{nvidia}
\mlsysauthor{Carole-Jean Wu}{meta}
\mlsysauthor{Christos Kozyrakis}{stanford}
\end{mlsysauthorlist}

\mlsysaffiliation{meta}{Meta, Menlo Park, California, USA}
\mlsysaffiliation{stanford}{Stanford University, Stanford, California, USA}
\mlsysaffiliation{nvidia}{NVIDIA, Santa Clara, California, USA}

\mlsyscorrespondingauthor{Mark Zhao}{myzhao@cs.stanford.edu}

\mlsyskeywords{Machine Learning, MLSys}

\vskip 0.3in
\begin{abstract}
We present \batchreuse (\textit{Rec}ommendation \textit{D}eduplication), a suite of end-to-end infrastructure optimizations across the Deep Learning Recommendation Model (DLRM) training pipeline.
\batchreuse addresses immense storage, preprocessing, and training overheads caused by feature duplication inherent in industry-scale DLRM training datasets.
Feature duplication arises because DLRM datasets are generated from interactions.
While each user \textit{session} can generate multiple training samples, many features' values do not change across these samples.
We demonstrate how \batchreuse exploits this property, end-to-end, across a deployed training pipeline.
\batchreuse optimizes data generation pipelines to decrease dataset storage and preprocessing resource demands and to maximize duplication within a training batch.
\batchreuse introduces a new tensor format, \textit{InverseKeyedJaggedTensors} (IKJTs), to deduplicate feature values in each batch.
We show how DLRM model architectures can leverage IKJTs to drastically increase training throughput.
\batchreuse improves the training and preprocessing throughput and storage efficiency by up to $2.48\times$, $1.79\times$, and $3.71\times$, respectively, in an industry-scale DLRM training system.
\end{abstract}

]

\printAffiliationsAndNotice{}  %

\section{Introduction}\label{sec:introduction}
Machine learning (ML) infrastructure is one of the most dominant components of industry-scale datacenters.
For example, ML consumes over $70\%$ of FLOPs at Google~\cite{computer22:patterson_carbon}.
To support the computational demands of ML, and especially training, companies such as Google~\cite{website:tpuv4-cloud}, Meta~\cite{isca22:mudigere_zionex, meta:rsc}, and AWS~\cite{website:aws_trainium} are deploying massive clusters consisting of tens of thousands of accelerators.

Deep learning recommendation model (DLRM) training is a principal industrial use-case for these clusters.
For example, DLRM training dominates ML capacity across Meta's fleet~\cite{arxiv:naumov_fb-dlt}.
This demand is driven by the ubiquity of DLRMs across industry, as they underlie critical services from Google~\cite{orsum22:rohan_factoryfloor, www20:li_search-ranking, recsys19:zhao_youtube}, Taobao~\cite{cikm18:ge_image-matters}, Meta~\cite{website:instgram-ranking, hpca18:hazelwood_applied-ml, acun:hpca21}, and others.
DLRM training clusters are fed by a data storage and ingestion (DSI) pipeline --- systems that generate, store, and preprocess training data --- which can demand more power consumption than what is required by the training accelerators (trainers) themselves~\cite{isca22:zhao_dsi}.
To train larger, more complex, and more accurate models, it is critical to improve the performance and efficiency of the \textit{end-to-end DLRM training pipeline}, from DSI to trainers.

To this end, this paper presents a suite of optimizations, called \batchreuse, spanning the DLRM training pipeline. %
\batchreuse exploits the inherent \textit{session-centric} nature of DLRM datasets.
DLRM training samples are generated from user interactions which query industrial recommendation models.
Each user's \textit{session} typically requires numerous inferences, and thus produces many training samples~\cite{survey21:wang_sbrs}.
However, the \textit{features} that largely compose each sample likely remain static throughout each session.
For example, an e-commerce DLRM dataset may contain a user feature representing the sequence of the last $N$ items added to a shopper's cart.
The e-commerce site may serve recommendations throughout a user's shopping session, but if the shopper does not add a new item, each of the session's samples will contain \textit{the same values} for that feature.

\begin{figure*}[t]
\centering
  \includegraphics[width=\linewidth]{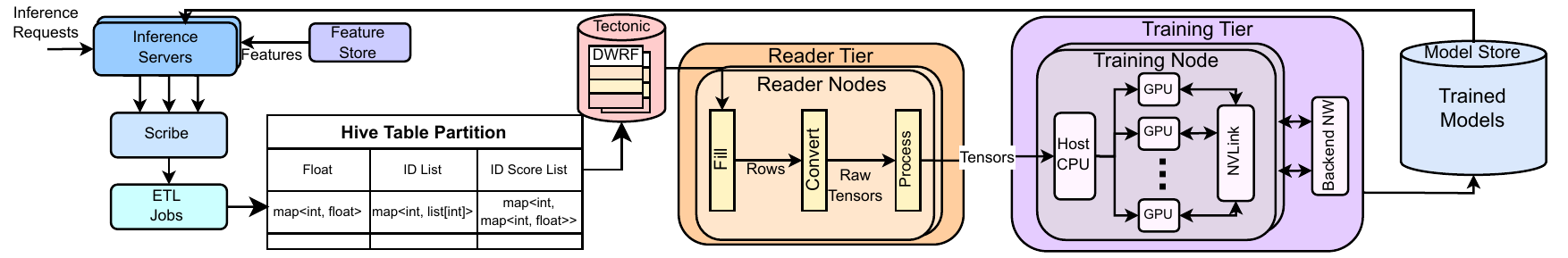}
    \vspace{-5mm}
  \caption{\small \revised{Industrial DLRM training pipeline composed of DSI and training services.}}
  \label{fig:infra_bd}
  \vspace{-3mm}
\end{figure*}

While prior work has mentioned feature duplication~\cite{cikm18:ge_image-matters, arxiv:gai_lsplm}, none has characterized its prevalence in industry-scale datasets nor provided solutions that optimize for it across the training pipeline.
Current pipelines spend considerable resources storing, preprocessing, and training over duplicate features.
These overheads constrain industry-scale training infrastructures from supporting larger datasets, longer features, and more complex modeling techniques (e.g., attention) that yield more accurate models~\cite{arxiv:ardalani_scaling,  recsys21:desouza_transformers4rec, transinfsyst21:fang_sequential, cikm19:li_tmall}.

We begin the paper with an in-depth characterization of how session-centricity generates significant feature value duplication in datasets used by an industrial DLRM training pipeline. %
Each session produces many samples (16.5 on average), and feature values are largely duplicated across the session's samples ($81.6-89.4\%$ on average).
\batchreuse addresses the significant storage, preprocessing, and training overheads caused by duplication throughout the DLRM training pipeline.

\batchreuse begins at data generation by sharding raw inference logs by session ID to improve compression ratios in \scribe~\cite{website:scribe}, a distributed message passing system.
These logs are ingested by ETL engines to produce training samples.
\batchreuse coalesces each session's samples within a training batch.
Not only does this reduce dataset sizes due to native compression, it also allows \batchreuse to convert each batch to a new tensor format during data reading, InverseKeyedJaggedTensors (IKJTs), that deduplicates feature values.

IKJTs require minimal resource overheads to generate and use for preprocessing and training.
Meanwhile, they allow readers, which preprocess data, and trainers to operate on deduplicated tensors, significantly reducing resource demands across the training pipeline.
We explore these benefits.
We present how DLRM architectures can leverage IKJTs to reduce GPU compute, network, and memory resource requirements --- improving training throughput and enabling more powerful modeling techniques.
In summary:

\begin{itemize}[leftmargin=*]
    \item We provide a characterization using petabyte-scale industrial DLRM datasets showing how feature duplication is inherent in DLRM training pipelines. We discuss the opportunities and challenges of deduplication. %
    \item We present necessary optimizations made in the data storage and ingestion pipeline to enable a novel tensor format, IKJTs, that deduplicates features in each training batch. %
    \item We show how IKJTs improve DLRM training throughput and resource utilization by eliminating redundant compute, memory, and network usage during training.
    \item We evaluate on industrial DLRMs. \batchreuse improves training and preprocessing throughput and storage efficiency by up to $2.48\times$, $1.79\times$, and $3.71\times$, respectively.
\end{itemize}

\section{Background}\label{sec:background}

Figure~\ref{fig:infra_bd} shows an end-to-end industrial training pipeline~\cite{isca22:zhao_dsi}, with DSI and training services.

\subsection{Data Storage and Ingestion}\label{subsec:background_dsi}
\noindent\textbf{Data Generation.}
Training data is continuously generated from deployed recommendation services. %
User-facing services request batches of inferences throughout a user's \textit{session}.
For each batch of requests, features corresponding to the user and potentially recommended items are retrieved from a feature store and are used as input to the DLRM to generate relevant predictions.
Since features continuously change, inference servers log features for each request to avoid data leakage~\cite{tkdd12:kaufman_leakage}.
Given predictions, user-facing services generate relevant impressions of items and log events (i.e., impression outcomes).
Logs are aggregated in \revised{\scribe, a global distributed messaging system~\cite{website:scribe}}.

Streaming and batch processing engines, such as Spark~\cite{nsdi12:zaharia_spark}, ingest data from \scribe.
These engines join raw features and events to produce labeled samples.
Training samples are subsequently landed into time partitioned (e.g., hourly) Hive tables~\cite{vldb09:thusoo_hive}.
To maintain data freshness, new table partitions are constantly landed and old partitions are deleted.

\noindent\textbf{Dataset Schema and Storage.}
Each training sample, corresponding to an impression and outcome, is stored as a structured row containing \textit{features} and \textit{labels}.
Features represent almost all of the bytes within a sample.
DLRMs require two types of features: dense and sparse.
Dense features represent continuous values, such as time, and are stored as a map from feature key to a float value.
Sparse features represent categorical values, such as item IDs, and are stored in map columns that map a feature key to its value, typically a variable-length list of item IDs.
\revised{Compared to dense features, sparse features require significantly more storage, preprocessing, and training resources across the DLRM training pipeline~\cite{isca22:zhao_dsi, arxiv:naumov_fb-dlt, asplos22:sethi_recshard}.}

Hive partitions are stored as columnar \revised{DWRF~\cite{isca22:zhao_dsi}} files similar in format to ORC~\cite{website:apache-orc}.
Files are composed of regions, called \textit{stripes}, that represent a small set of rows.
Within a stripe, rows are stored as columnar streams.
Feature columns are first flattened (i.e., each feature key becomes a separate column).
Values for each flattened column (e.g., ID lists) are then encoded and compressed into streams.
Files are stored in \revised{\tectonic, an exabyte-scale distributed filesystem~\cite{fast21:pan_tectonic}}.

\noindent\textbf{Data Reading and Preprocessing.}
Each DLRM training job specifies its \textit{dataset} (table partitions) and \textit{preprocessing} needs via a DataLoader specification~\cite{website:pytorch-dataloader}.
A \textit{reader tier}, composed of stateless \textit{readers}, is launched for each job.
Readers scan through the dataset partitions and preprocess rows into tensors.
Each reader \textit{fills} a batch of rows by reading files from \tectonic and \textit{converts} the batch into raw tensors.
The reader then \textit{processes} the tensors according to the job's specifications by applying transformations such as normalization and hashing.
Preprocessed tensors are sent to trainers.
The number of readers for each job is scaled to meet trainers' ingestion bandwidth demands.

\subsection{DLRM Training at Scale}\label{subsec:background_training}
\begin{figure}[t]
  \centering
  \includegraphics[width=\linewidth]{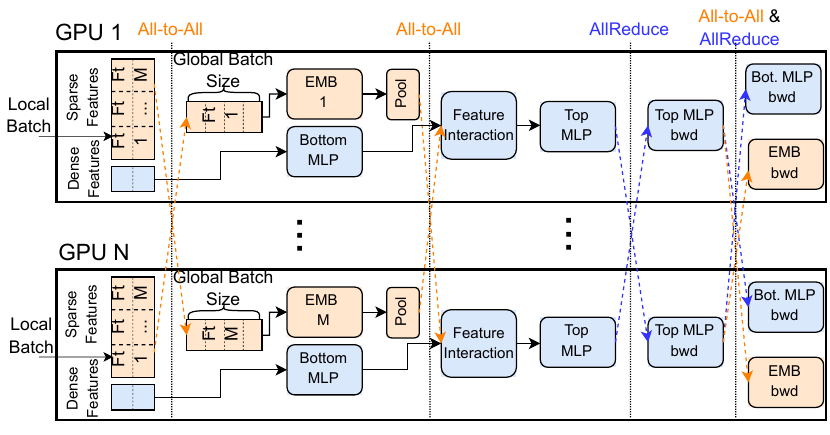} 
    \vspace{-5mm}
  \caption{\small How DLRMs are synchronously trained using hybrid parallelism across multiple GPUs. Dashed lines show collective communication across \textcolor{orange}{model-parallel} and \textcolor{blue}{data-parallel} modules.}
  \label{fig:dlrm_background}
  \vspace{-3mm}
\end{figure}

Figure~\ref{fig:dlrm_background} shows how a typical DLRM~\cite{arxiv:naumov_dlrm} is synchronously trained across multiple GPUs.
DLRMs primarily consist of multilayer perceptrons (MLPs) and embedding tables (EMBs) composed into three main architectural components.
EMBs ingest sparse feature lists and produce a dense activation vector for each list element. %
A pooling function (e.g., average, sum, or max) aggregates activations for each sparse feature.
Meanwhile, a bottom MLP transforms dense features into a dense representation with the same dimensionality as embedding vectors.
An interaction layer explicitly computes second-order interactions across dense and sparse features (e.g., via pairwise dot product).
A top MLP and sigmoid processes the result to produce a probability output (e.g., click-through rate).

DLRMs are trained using hybrid parallelism across multiple GPUs.
MLPs are copied across GPUs in a distributed data parallel (DDP) fashion, while EMBs are sharded across GPUs via distributed model parallelism (DMP) due to their large size.
During each training iteration, each GPU ingests a local batch from the reader tier.
A sparse data distribution (SDD) step first aggregates the appropriate feature values, across all local batches, to the corresponding GPU using an all-to-all collective~\cite{nvidia:nccl} across all GPUs.
After the EMB lookup and pooling, another all-to-all distributes embedding vectors back to their original GPUs, as feature interaction and the top MLP is data parallel.
After calculating the loss, an all-reduce aggregates gradients to update MLPs.
Similarly, an all-to-all synchronizes EMB parameter updates during the backward pass.
Thus, the iteration time is determined by both the per-GPU compute and memory bandwidth resources (for MLPs, interactions, pooling, and EMB lookups), as well as the backend network bandwidth and latency (for collective communications).

\noindent\textbf{Scaling Systems for DLRMs.}
Improving DLRM accuracy necessitates systems efficiency and throughput optimizations across the training pipeline.
For example, \cite{arxiv:ardalani_scaling} showed that data scaling significantly improves DLRM performance.
Supporting growing dataset volumes requires not only more efficient storage, but also improved reader and trainer throughputs to complete training within a reasonable amount of time.
Meanwhile, recent DLRM architectures focus on capturing users' long-term interests via a sequential history of interactions~\cite{kdd19:pi_uic, cikm19:li_tmall, dlpkdd19:chen_behavior-sequence-transformer}.
These architectures use long \textit{sequence features} and attention mechanisms, such as transformers~\cite{neurips17:vaswani_attention}, to pool embeddings across many sequence features.
They demand significant GPU compute, memory, and network resources.
Thus, optimizing for DSI and training performance and efficiency is increasingly urgent as resource demands continue to grow.

\section{Understanding Data Reuse}\label{sec:characterization}

To understand the opportunity for \batchreuse, we explore the prevalence of duplication within industry-scale datasets.
\revised{We focus on sparse features because they \textit{a)} are prone to duplication as we characterize next, and \textit{b)} demand significantly more training pipeline resources than dense features (Section~\ref{sec:background}) and thus present a more attractive optimization target.}

Duplication arises because \revised{sparse} user features rarely change across impressions within a session\footnote{A session is a set of user impressions in a fixed time window.}.
For example, consider social media features $f_{like}$ and $f_{share}$, which contain a user's last N posts they liked and shared, respectively.
During a session, a user may view multiple posts (impressions).
While each impression may generate a training sample, $f_{like}$ and $f_{share}$ will be exactly the same across samples if the user did not like or share a post.
Some features, such as $f_{share}$, may rarely change, and even if they do, their lists will be shifted with most elements being the same.
In summary, duplication depends on \textit{the number of samples per session}, and is a per-feature property depending on \textit{how often each feature's value changes across samples}. %

\begin{figure}[t]
  \centering
  \includegraphics[width=\linewidth]{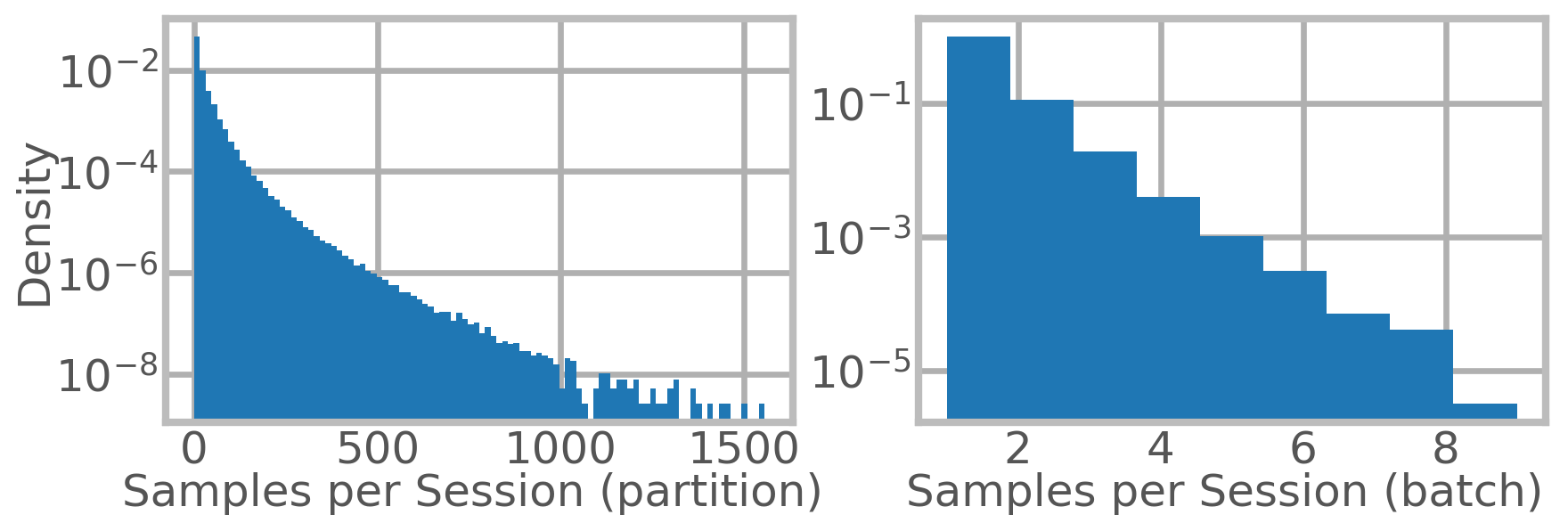} 
    \vspace{-5mm}
  \caption{\small Histogram of the number of samples per session within an hourly partition (left) and 4096 batch (right) from the partition.}
  \label{fig:session_histogram}
   \vspace{-3mm}
\end{figure}

\noindent\textbf{Each session generates multiple training samples, but they are distributed across training batches.}
Figure~\ref{fig:session_histogram} shows a histogram of the number of training samples generated by each session in a representative, \revised{O(100PB)-scale industrial DLRM dataset}.
We show graphs for an hourly table partition and a batch of 4096 samples from the partition.

We observe that, on average, each session generates $16.5$ samples within an hourly partition with a significant tail of over 1000 samples per session.
While this shows that there is significant opportunity for deduplication, the training pipeline operates on a relatively small subset of training samples at a time (e.g., file splits and training batches).
Deduplicating a session's samples requires us to co-locate the samples closely together within the table.

However, the data generation infrastructure typically orders samples based on inference time.
The large volume of inference requests across services naturally interleaves samples across sessions.
Figure~\ref{fig:session_histogram} demonstrates how within each batch of 4096 samples, this interleaving results in only $1.15$ samples per session on average.
Optimizations must be co-designed alongside data generation infrastructure to fully coalesce and deduplicate samples within each session.%

\begin{figure}[t]
  \centering
  \includegraphics[width=\linewidth]{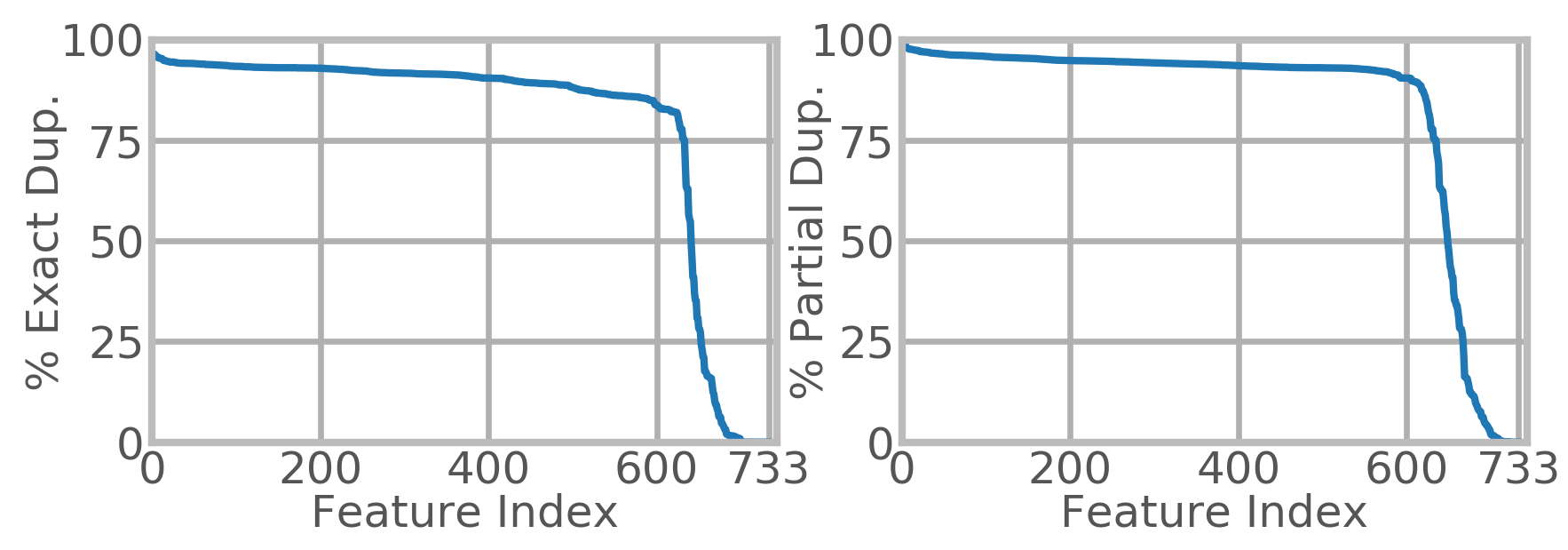} 
    \vspace{-5mm}
  \caption{\small Percent of exact (left) and partial (right) duplicate values across sparse features within an hourly partition.}
  \label{fig:matches}
  \vspace{-3mm}
\end{figure}

\noindent\textbf{Within a session's samples, there is a large amount of exact and partial feature duplication.}
It is also important to validate that feature values (i.e., lists) themselves are largely duplicated across these samples.
We first quantify the amount of exact duplicate feature values across samples. %
Specifically, for given feature $x$, we analyze the percent of samples within the hourly partition that contained exactly the same list as another sample from the same session within the partition.
For example, if feature $x$ is never updated across $16.5$ samples per session, we would expect a maximum of $15.5/16.5 = 93.9\%$ of exact duplicates in the partition.
Figure~\ref{fig:matches} shows this result across 733 sparse features in the hourly partition.
We observe that on average across all features, $80.0\%$ of feature values are an exact duplicate. %
This validates our assumption that many DLRM features are not updated across a session's samples.

\revised{Specifically, DLRM sparse features largely reflect either user or item traits.
User sparse features (e.g., last N liked item IDs) are largely duplicated across a user's samples.
Item features (e.g., the item ID that is evaluated for recommendation) are less duplicated since many different items are ranked in a given session.
Figure~\ref{fig:matches} shows this distinction -- user features comprise the vast majority of dataset volume and accordingly represent the large subset of features with high duplication.
Meanwhile, item features exhibit less duplication, representing the subset of features right of the knee.
We expect increased reliance on user features, and thus higher feature duplication, since recommender systems are increasingly focusing on larger user interaction history features compared to item metadata (Section~\ref{subsec:background_training}).}

For highly-duplicated user features, even if its values change across samples, we expect the majority of its list IDs to remain the same.
We thus repeated the analysis on an individual list ID basis.
For example, suppose feature $x$ contained $100$ IDs across $2$ training samples.
$x$ may be updated by appending a new ID and shifting its list by one, resulting in $99/200 = 49.5\%$ partial duplication.
Figure~\ref{fig:matches} shows how on average across all features, $83.9\%$ of feature values within each feature list are duplicated.
Many non-exact duplicate samples within the session contain partial duplicates.

Finally, it is important to note that \textit{not all feature lists have the same length}.
To understand how many bytes are duplicated, we weigh each feature in our prior analysis by its respective average length.
We find that $81.6\%$ and $89.4\%$ of all IDs in feature values (i.e., bytes) are exact and partial duplicates, respectively, suggesting that longer features have slightly more exact and partial duplicates.

\begin{table*}[t]
  \centering
  \caption{\small Overview of \batchreuse optimizations made throughout the industry-scale training pipeline.}
    \label{tbl:optimizations}
  \footnotesize
  \begin{tabular}{p{2.5cm}|p{2cm}|p{11.25cm}}
  \toprule
  \textbf{Optimization} &  \textbf{Target System} & \textbf{Benefit} \\ \midrule
  \textbf{O1}: Log Sharding ($\S$\ref{sec:dsi_generation})
  & \scribe 
  & Improves black-box compression ratios to reduce \scribe network RX/TX and storage demands. \\ \hline
  \textbf{O2}: Cluster by Session ($\S$\ref{sec:dsi_generation})
  & ETL
  & Session sample co-location enables readers/trainers to exploit duplicate features. Improves file compression ratios, reducing storage and read IOPS demands. \\ \hline
  \textbf{O3}: Inverse KJTs ($\S$\ref{sec:dsi_ikjt})
  & Readers 
  & New tensor encoding allows downstream preprocessing/training operations to use deduplicated features, enabling significant resource savings. \\ \hline
  \textbf{O4}: Deduplicated Preproc. ($\S$\ref{sec:dsi_preproc})
  & Readers
  & IKJT preprocessing modules reduce preprocessing compute demands. Deduplicated outputs require less NW bandwidth between readers and trainers. \\ \hline
  \textbf{O5}: Deduplicated EMB ($\S$\ref{sec:training})
  & Trainers 
  & Reduced per-iteration trainer compute/memory/NW demands by deduplicating EMB features, lookups, and activations. \\ \hline 
  \textbf{O6}: JaggedIndex-Select ($\S$\ref{sec:training})
  & Trainers
  & Reduced memory copy overheads by enabling index select without first converting jagged tensors to a dense representation. \\ \hline 
  \textbf{O7}: Deduplicated Compute ($\S$\ref{sec:training})
  & Trainers
  & Reduced compute for sparse feature modules (especially attention pooling) by allowing them to operate on deduplicated tensors.
  \\ \bottomrule
  \end{tabular}
  \vspace{-5mm}
\end{table*}

\noindent\textbf{Summary.}
The vast majority of feature values are duplicated within the industry-scale DLRM dataset.
While significant deduplication opportunities exist, they require each session's samples to be co-located within training batches.
Thus, trainer-only solutions are insufficient --- optimizations must be co-designed across the end-to-end training pipeline.

\section{\batchreuse in Data Storage and Ingestion}\label{sec:dsi}

\batchreuse implements these optimizations, summarized in Table~\ref{tbl:optimizations}, throughout the industrial pipeline shown in Figure~\ref{fig:infra_bd}.

\subsection{Data Generation and Storage}\label{sec:dsi_generation}
\noindent\textbf{Log Sharding.}
\revised{\scribe} is a message passing service which logically aggregates and buffers raw logs from each inference server.
To load balance, \revised{\scribe} consistently hashes the message and routes each to a \textit{shard} on a physical storage node, which buffers and compresses messages in memory and on disk.
Unfortunately, the default hashing configuration distributes logs for each session randomly across shards.
\batchreuse configures \revised{\scribe} to instead \textit{use session IDs as the shard key}, improving the ``compressibility'' of data within each shard.
Thus, we can both reduce the number of \revised{\scribe} storage nodes and the amount of network bandwidth needed for downstream ETL jobs to ingest logs.

\noindent\textbf{Clustering by Session.}
While improved sharding increases the locality of a session's logs, it does not guarantee that a session's training samples are adjacent within the dataset.
This grouping is needed for downstream systems to deduplicate features. %
Thus, \batchreuse adds a data generation ETL job, which clusters partitions by session ID and sorts by log timestamp.
As with \scribe, we also expect two direct benefits from ETL clustering.
First, each file's stripes are compressed using black-box compression, e.g. zstd~\cite{website:zstd}.
Ensuring that each stripe contains multiple rows for a given session increases compression ratios and thus reduces dataset storage requirements.
Secondly, smaller files also reduce compute and network resources needed to read samples during online preprocessing.

\subsection{Tensor Encoding for Deduplication}\label{sec:dsi_ikjt}
\begin{figure}[t]
  \centering
  \includegraphics[width=\linewidth]{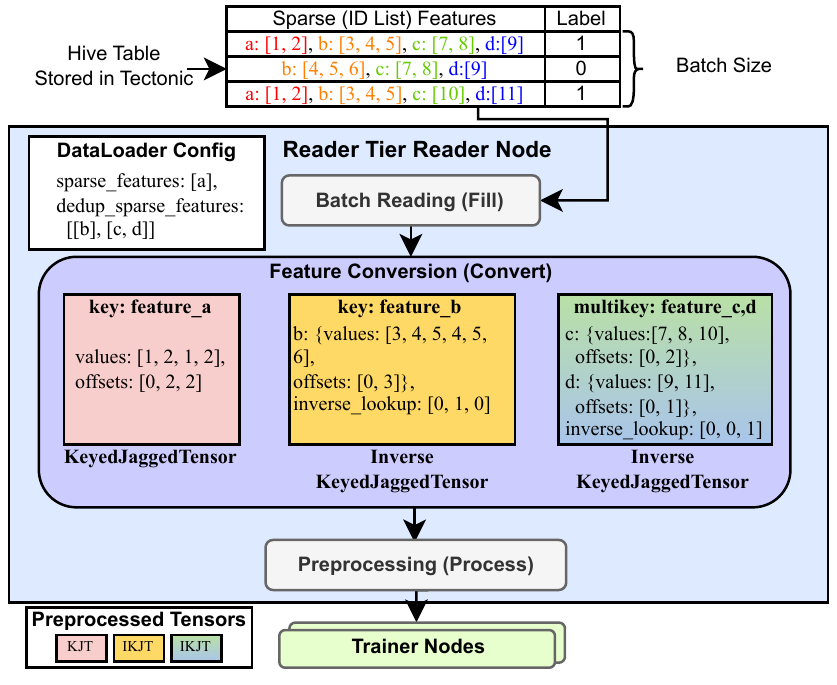} 
    \vspace{-3mm}
  \caption{\small Each reader node extracts mini-batches from storage. Batches are converted to tensors, preprocessed, and sent to trainers.}
  \label{fig:ikjt}
\end{figure}

Figure~\ref{fig:ikjt} shows how reader nodes generate preprocessed tensors for each training job.
Each reader reads batches of samples, converts rows to tensors, and preprocesses tensors.

A \textit{Feature Conversion} step copies data from raw batches of rows, read into memory as byte arrays, into structured tensors.
The typical tensor format used for sparse features is a KeyedJaggedTensor (KJT)~\cite{website:torchrec-sparse}.
A KJT maps a key (i.e., the feature key) to a JaggedTensor --- a tensor with a jagged dimension (i.e., different length slices). %
For example, Figure~\ref{fig:ikjt} shows how a batch of 3 rows for feature $a$ is transformed into a KJT with two slices representing the feature's $values$ and $offsets$ for the batch.
The $offsets$ slice has an entry for each row, with $offsets[i]$ pointing to the starting index in the $values$ slice for row $i$.
\revised{The length of the feature for row $i$ is calculated from $offsets[i+1] - offsets[i]$ (or $|values| - offsets[i]$ for the last row).}
In the example, feature $a$ has $2\times$ duplication in the $values$ slice as rows 0 and 2 both contain $[1, 2]$. %

\noindent\textbf{InverseKeyedJaggedTensor.} To deduplicate feature values, \batchreuse introduces a new \textit{inverse} KJT format. %
Figure~\ref{fig:ikjt} shows how ML engineers can specify a \texttt{dedup\_sparse\_features} field in the PyTorch DataLoader, which is a \texttt{List[List[featureKey]]}, containing lists of \textit{feature groups} to deduplicate.
\batchreuse deduplicates each feature group to an InverseKeyedJaggedTensor (IKJT) during feature conversion.
Of course, users can still generate KJTs for features not exhibiting high duplication.

\batchreuse deduplicates features by detecting and avoiding duplicate copies during feature conversion.
An IKJT instead adds an additional $inverse\_lookup$ slice, where $inverse\_lookup[i]$ points to the respective entry in the deduplicated $offsets$ slice for row $i$ in the batch.
$offsets$ encodes the $values$ slice as before.
In our example, feature $b$ contains duplicate values for rows 0 and 2.
Thus $inverse\_lookup[0] == inverse\_lookup[2]$, with both pointing to $offset[0]$ which encodes the duplicate values $[3, 4, 5]$.
IKJTs avoid storing a second copy of $[3, 4, 5]$ in the $values$ slice.
Since exact matches are the vast majority of duplication (Section~\ref{sec:characterization}), we focus on deduplicating exact matches and discuss supporting partial matches in Section~\ref{sec:discussion}.

\noindent\textbf{Grouped IKJTs.}
Users can deduplicate multiple features within a single IKJT.
Grouped IKJTs are designed for features updated synchronously across samples and thus share $inverse\_lookup$ values. %
For example, an e-commerce model may use two features which track the item ID and seller ID for items added to a user's cart.
Since both features track the same item sequence, they are both updated at the same time (i.e., when a new item is added).
As we explore in Section~\ref{sec:training}, grouped IKJTs are designed to enable additional optimizations during each training iteration. %

In Figure~\ref{fig:ikjt}, features $c$ and $d$ are deduplicated as a group.
For both features, rows 0 and 1 are duplicates.
Thus, \batchreuse uses a common $inverse\_lookup$ to reference the $offsets$ slice for both features, even if their respective offsets or values slices are different.
For example, $inverse\_lookup[0]$ will map to $[7, 8]$ for feature $c$ and $[9]$ for feature $d$.
In the event that grouped feature values are not synchronously updated across samples, we will not deduplicate the corresponding unsynchronized rows to ensure that the $inverse\_lookup$ invariant is maintained.

\noindent\textbf{Using IKJTs.}
Not all features may be worth deduplicating.
To understand the value of deduplication, we use the following analytical model for a feature $f$.
$S$ is the average number of samples per session.
$B$ is the batch size.
$d(f)$ is the probability that the $f$'s value will remain the same across adjacent rows.
$l(f)$ is the average length of $f$. %
\begin{align*}
&DedupeLen(f) = l(f)*B*(1 - (S-1) * S^{-1} * d(f)) \\
&DedupeFactor(f) = l(f) * B / DedupLen(f) 
\end{align*}
Specifically, $DedupeLen(f)$ expresses the size of the $values$ slice after deduplicating $f$ for each training batch. %
The deduplication factor, $DedupeFactor(f)$, is calculated as the ratio of the original $values$ slice length to $DedupeLen(f)$. %
For example, suppose $B = S = 3$, $l(b) = 3$, and $d(b) = 0.5$ for feature $b$ in Figure~\ref{fig:ikjt}.
Deduplicating $b$ results in $DedupeLen(b) = 6$ and $DedupeFactor(b) = 1.5$.
The total amount of feature values deduplicated increases with higher $S$, $l(f)$, and $d(f)$, which aligns (i.e., increases) with data scaling trends (Section~\ref{subsec:background_training}).

While $inverse\_lookup$ and $offsets$ requires more elements than $offsets$ alone (up to $B$), the overhead is negligible as for most features $l(f)*B >> B$.
Furthermore, as we discuss in Section~\ref{sec:training}, because only $values$ and $offsets$ tensors are communicated across GPUs, IKJTs strictly decrease over-the-network tensor sizes during training.
Finally, $DedupeFactor(f)$ provides an initial guidance on the impact of deduplicating $f$.
We typically deduplicate features with $DedupeFactor(f) > 1.5$.
However, the actual performance benefit depends on how well readers and trainers can use IKJTs, as we explore next and discuss in Section~\ref{sec:discussion}.

\subsection{Preprocessing over IKJTs.}\label{sec:dsi_preproc}
After feature conversion, each reader node preprocesses tensors using a set of user-provided TorchScript modules.
If a user deduplicates a feature, we automatically add a wrapper that transparently supports preprocessing over IKJTs.
Since the original function used KJTs, the wrapper simply provides the $offsets$ and $values$ slices from the IKJT held in memory, saving significant compute resources by avoiding preprocessing duplicate values.
Deduplicated preprocessing functions also output IKJTs.
This reduces network bandwidth requirements between reader and trainer nodes and allows trainers to further leverage IKJTs.

\section{\batchreuse in Training}\label{sec:training}
\begin{figure}[t]
  \centering
  \includegraphics[width=\linewidth]{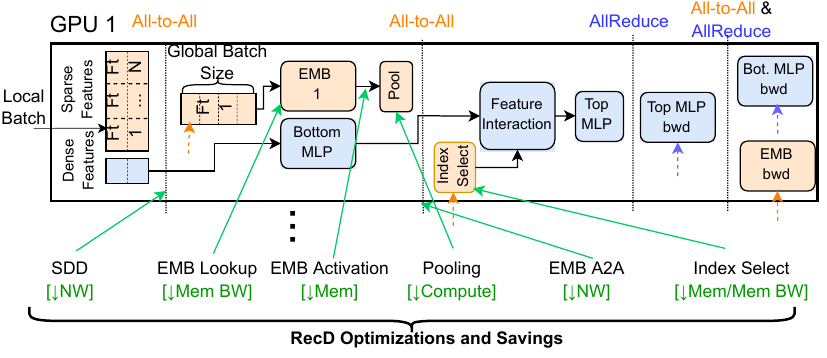} 
    \vspace{-3mm}
  \caption{\small Overview of a training iteration, with green arrows showing where \batchreuse optimizations save trainer resources. Dashed arrows show \textcolor{orange}{all-to-all} and \textcolor{blue}{all-reduce} communication.}
  \label{fig:trainer_opts}
\end{figure}

Building on our newly proposed IKJT tensor format, we design a series of \batchreuse PyTorch modules as direct replacements for DLRM embedding and pooling operations.
The IKJT format generated by readers enables a host of optimizations at the trainer, summarized in Table~\ref{tbl:optimizations}.
As shown in Figure~\ref{fig:trainer_opts}, these modules operate on \textit{deduplicated} (i.e., IKJT) tensors during the forward pass, reducing resources spent operating over duplicate sparse feature values.

\noindent\textbf{Sparse Data Distribution.}
After receiving a batch of samples, each GPU executes a sparse data distribution (SDD) step.
Using an all-to-all collective, SDD coalesces a global batch only containing the respective features corresponding to each GPU's model-parallel EMBs.
Previously, KJTs required sending significant amounts of duplicate feature values over the network.
With \batchreuse, deduplicated IKJT $value$ and $offset$ slices are sent instead ($inverse\_lookup$ slices are kept local).
\batchreuse thus reduces the amount of bytes distributed during SDD by $DedupeFactor(f)$ for each feature $f$.
Since SDD runs before any embedding lookups, reducing the amount of data over the network in each iteration directly improves training throughput.

\noindent\textbf{EMB Lookups.}
After SDD, each trainer needs to translate every feature value in the KJT into an embedding by performing a lookup in each EMB.
By using IKJTs, the length of the $values$ slice is reduced by $DedupeFactor(f)$, reducing the overall number of EMB lookups we need to perform in each iteration and thus required memory bandwidth.

\noindent\textbf{EMB Inputs and Activations.}
Each GPU also needs to allocate significant dynamic memory to store the feature inputs and EMB activations of each sparse value.
This is especially true of long length sequence models.
For example, a single feature $f$ with $l(f) = 1000$, $B = 4096$, and an EMB dimension of 128 would require $4096*1000*128*4B \approx 2GB$ of GPU memory to store activations.
By performing lookups using IKJT values, we directly reduce the amount of dynamic GPU memory required by $DedupeFactor(f)$.

\noindent\textbf{Deduplicated Pooling.}
DLRMs use a set of pooling modules (e.g., sum, avg.) that operate on the EMB activations prior to feature interaction.
Recent trends have motivated more complex pooling modules, such as transformers and other attention mechanisms~\cite{kdd19:pi_uic, cikm19:li_tmall}, which operate over multiple long-length sequence features. %
These modules require significant GPU resources.

To reduce the computational and memory overhead for these sequential pooling modules, \batchreuse allows users to run compute modules with IKJTs as inputs.
Specifically, by ensuring that the $inverse\_lookup$ slice is shared across all features within an IKJT, we can deduplicate compute by simply operating on the deduplicated $values$ and $offsets$.
For example, assume a module element-wise sums values for each row across features $c$ and $d$ in the example in Figure~\ref{fig:ikjt}.
Using KJTs, the GPU computes $[7+8+9, 7+8+9, 10+11] = [24, 24, 21]$.
With IKJTs, we instead compute $[7+8+9, 10+11] = [24, 21]$ and simply use the shared $inverse\_lookup$ to expand the output to $[24, 24, 21]$.
By applying this technique to expensive attention pooling modules, we reduce the compute demand by $DedupeFactor(f)$ for each sequence feature $f$.

\noindent\textbf{Deduplicated EMB.}
Since the output of pooling layers are still in the IKJT format, we can get more network savings during the all-to-all that broadcasts pooled embeddings back to each GPU for feature interaction.

\noindent\textbf{Jagged Index Select.}
Before feature interaction, IKJTs must be converted back to a KJT to be interacted with other non-deduplicated features.
We use \texttt{torch.index\_select} to perform this conversion.
Prior to \batchreuse, \texttt{index\_select} could only operate on dense, not jagged tensors.
We needed to first convert jagged tensors into dense tensors (e.g., via padding), incurring large memory overheads.
We implemented a jagged \texttt{index\_select} to operate over jagged tensors, eliminating this overhead.

\noindent\textbf{Summary.}
As summarized in Figure~\ref{fig:trainer_opts}, IKJTs enable a host of GPU network, memory, memory bandwidth, and compute optimizations during training.
These optimizations improve training throughput and reduce GPU resource demands, allowing us to train more complex models at a faster rate.

\section{Evaluation}\label{sec:evaluation}

\subsection{End-to-end Performance Optimizations}
\batchreuse improves the performance and efficiency of the entire training pipeline, including storage, readers, and trainers.
To study each component, we used three representative industrial DLRMs, $RM_{1}$, $RM_{2}$, and $RM_{3}$, \revised{designed around the core DLRM architecture~\cite{arxiv:naumov_dlrm}.
$RM_1$, $RM_2$, and $RM_3$ contain $O(10^9)$, $O(100^9)$, and $O(100^9)$ parameters with $O(10GB)$, $O(100GB)$, and $O(100GB)$ of embedding tables, respectively.
Embedding dimensions range from 64-1024 across each $RM$.}

\revised{We evaluate on a trainer tier consisting of ZionEX training nodes~\cite{isca22:mudigere_zionex}.
Each ZionEX node contains 8 NVIDIA A100 GPUs with a total of 320 GB HBM and 12.4 TB/s of memory bandwidth.
Intra-node communication across GPUs occurs via NVLink.
Each GPU is equipped with a 200 Gbps RoCE NIC for inter-node communication over a dedicated RoCE backend network.
Input data is supplied by 4 host CPU sockets, each with a 100 Gbps NIC that ingests data from a tier of DPP~\cite{isca22:zhao_dsi} readers.
Each reader is a general-purpose, x86 CPU server with 18 cores, a 12.5 Gbps NIC, and 64 GB of memory.
Readers read data from each $RM$'s respective $O(100PB)$ industrial dataset stored within the \tectonic file system~\cite{fast21:pan_tectonic}.}

For each $RM$, we used the default baseline configuration, with $RM_1$, $RM_2$, and $RM_3$ using a batch size of 2048, 2048, and 1152, and 48, 48, and 64 GPUs, respectively.
We then enabled the full suite of \batchreuse optimizations for each $RM$, which allowed us to increase the batch size for $RM_1$ and $RM_3$ to 6144 and 2048, respectively.
For $RM_2$, we could not substantially increase batch size beyond 2048.
For $RM_1$, we deduplicated 16 sequence features in 5 groups. 
$RM_2$ and $RM_3$ deduplicated 6 and 11 sequence features, respectively, in one group. 
Each $RM$ also deduplicated an additional $\approx100$ features that were element-wise (e.g., sum, max) pooled.
$DedupeFactor$ was $\approx4 - 15$ for deduplicated features.
We used a clustered table for \batchreuse models containing the same data as the baseline table.
We kept all other hyper parameters the same and scaled the number of readers to provide sufficient throughput to avoid data stalls in all configurations.

\begin{figure}[t]
\centering
  \includegraphics[width=\linewidth]{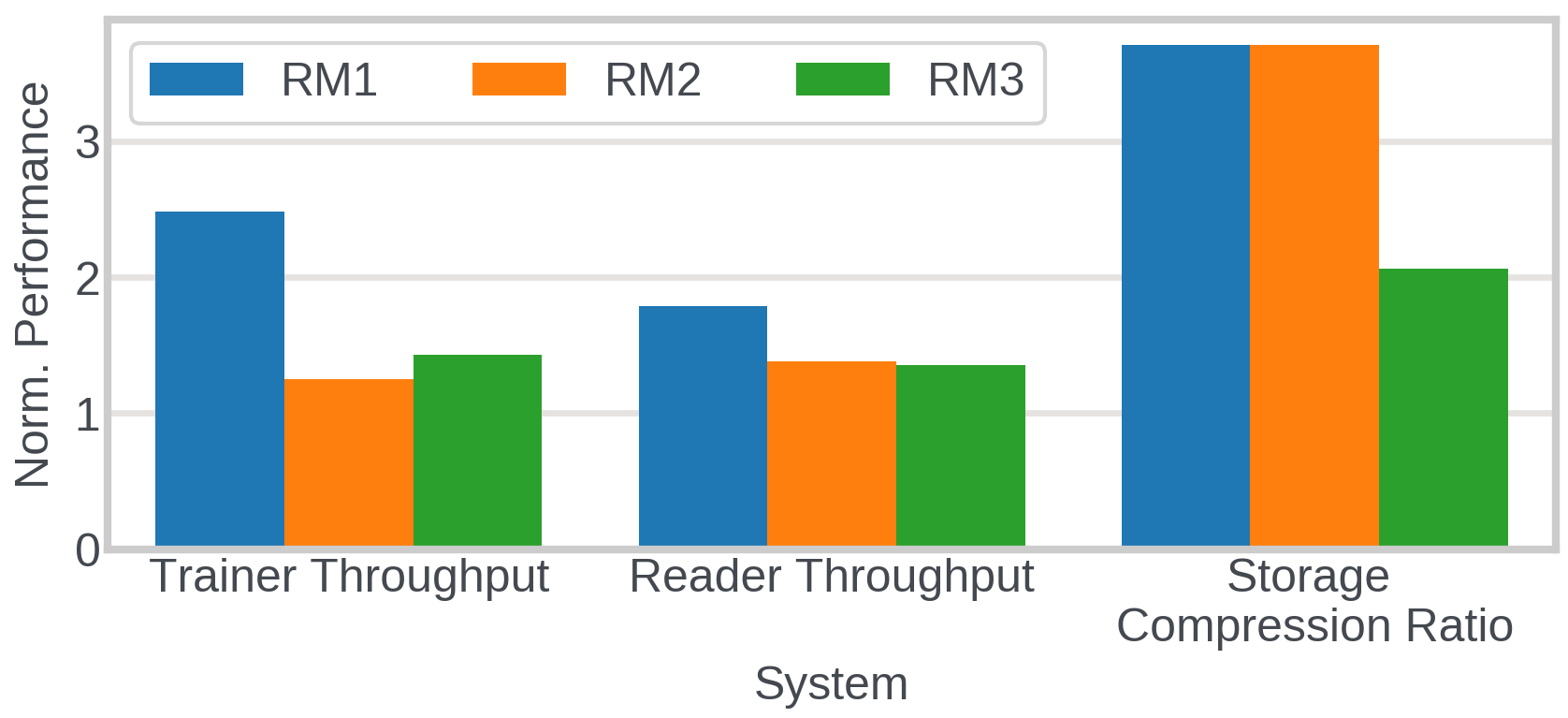}
    \vspace{-3mm}
  \caption{\small Trainer, reader, and storage performance using \batchreuse across three representative $RMs$, normalized to their baselines. } %
  \label{fig:model_eval}
\end{figure}

Figure~\ref{fig:model_eval} shows how trainer throughput, reader throughput, and storage compression ratio improved with respect to the baseline for each $RM$.
\revised{Trainer throughput is the total samples per second processed by all trainers.
Since we scale the number of readers based on trainer throughput, we report the samples per second processed on average by each reader.
Finally, we report the compression ratio of the clustered table's \tectonic files relative to the baseline table.}
\batchreuse improved trainer throughput by $2.48\times$, $1.25\times$, and $1.43\times$, significantly decreasing training job latencies. 
Similarly, each reader processed samples $1.79\times$, $1.38\times$, and $1.36\times$ faster, \revised{reducing the number of readers needed for each training job by the same amount.}
We explore in Section~\ref{sec:trainer_eval} and \ref{sec:reader_eval} why $RM_1$'s increased use of sequence features allowed \batchreuse to further increase trainer and reader throughput, respectively, compared to $RM_2$ and $RM_3$.
Clustered tables improved the compression ratio by $3.71\times$ ($RM_1$ and $RM_2$ used the same table) and $2.06\times$, \revised{directly improving storage efficiency by reducing the number of storage nodes needed to store and serve each $RM$'s dataset.}
$RM_1$ and $RM_2$'s table exhibited higher samples per session than $RM_3$'s table, leading to a larger increase in compression after co-locating each session's samples within a file stripe.
Finally, we increased the compression ratio at \scribe from $1.50\times$ to $2.25\times$ by sharding logs by session ID.

\subsection{Why does \batchreuse improve trainer throughput?}\label{sec:trainer_eval}
\begin{figure}[t]
\centering
  \includegraphics[width=\linewidth]{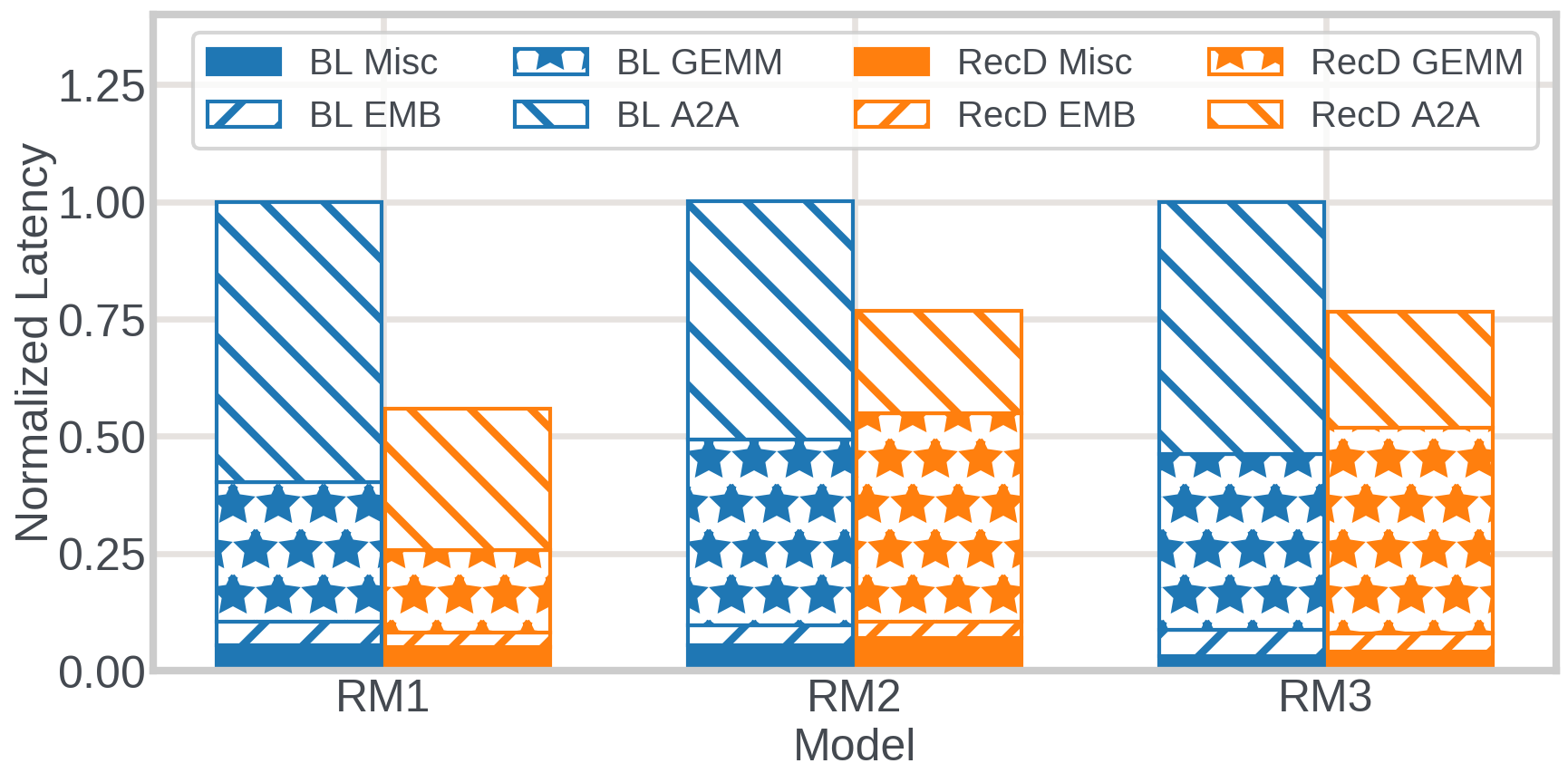}
    \vspace{-3mm}
  \caption{\small Breakdown of trainer iteration latency across $RM$s, normalized to each $RM$'s baseline latency using the same batch size.}
  \label{fig:trainer_eval}
\end{figure}

\noindent\textbf{Iteration Breakdown.}
To understand where \batchreuse improves training throughput, we ran a training job for each $RM$ \textit{using the same batch size as the baseline}.
Figure~\ref{fig:trainer_eval} shows a breakdown of the iteration latency of the \batchreuse training job normalized to the baseline iteration latency, averaged across all GPUs.
Specifically, we show \textit{exposed} latency (i.e., non-overlapping compute/communication), broken down by GPU time spent on EMB lookups, compute (GEMM), all-to-all communication (A2A), and other miscellaneous operations (e.g., all-reduce and reduce-scatter).

First, we observe that \batchreuse halves exposed A2A communication across all $RM$s.
A2A is a significant component of each training iteration.
\batchreuse significantly improves training throughput by reducing the amount of over-the-network bytes via IKJTs.
Thus one reason for $RM_1$'s larger training throughput gains is because it exposes more communication by using more sequence features that \batchreuse optimizes for.

The second reason is because $RM_1$ uses expensive transformers to pool EMB activations for several user sequence features; \batchreuse deduplicated the compute for these transformers by grouping each transformer's features together using IKJTs.
This is evidenced by an additional reduction ($12\%$ of iteration latency) in the amount of time spent in GEMMs for $RM_1$.
Meanwhile, $RM_2$ and $RM_3$ saw slight increases in exposed $GEMM$ time.
This is because less of it was hidden as \batchreuse reduced A2A latencies, as well as a slight increase due to the additional \texttt{index\_select}. %
We also observe a small improvement across $RM$s due to faster EMB lookups ($1-2\%$ of iteration latency) by eliminating redundant lookups.
While this did not significantly improve trainer throughput given the same batch size, using fewer EMB activations allowed us to increase batch size to improve trainer throughput, as we study next.

Finally, Figure~\ref{fig:trainer_eval} shows why translating $DedupeFactor$ to throughput gains is challenging.
$RM_1$ and $RM_2$ used the same table and features with similar $DedupeFactor$s. %
However, \batchreuse reduced $RM_1$'s iteration time by $44\%$ compared to $23\%$ for $RM_2$ due to the differences in model architectures and exposed compute/communication cycles as discussed above.
We discuss observations on how ML engineers choose which features to deduplicate in Section~\ref{sec:discussion}.

\begin{figure}[t]
\centering
  \includegraphics[width=\linewidth]{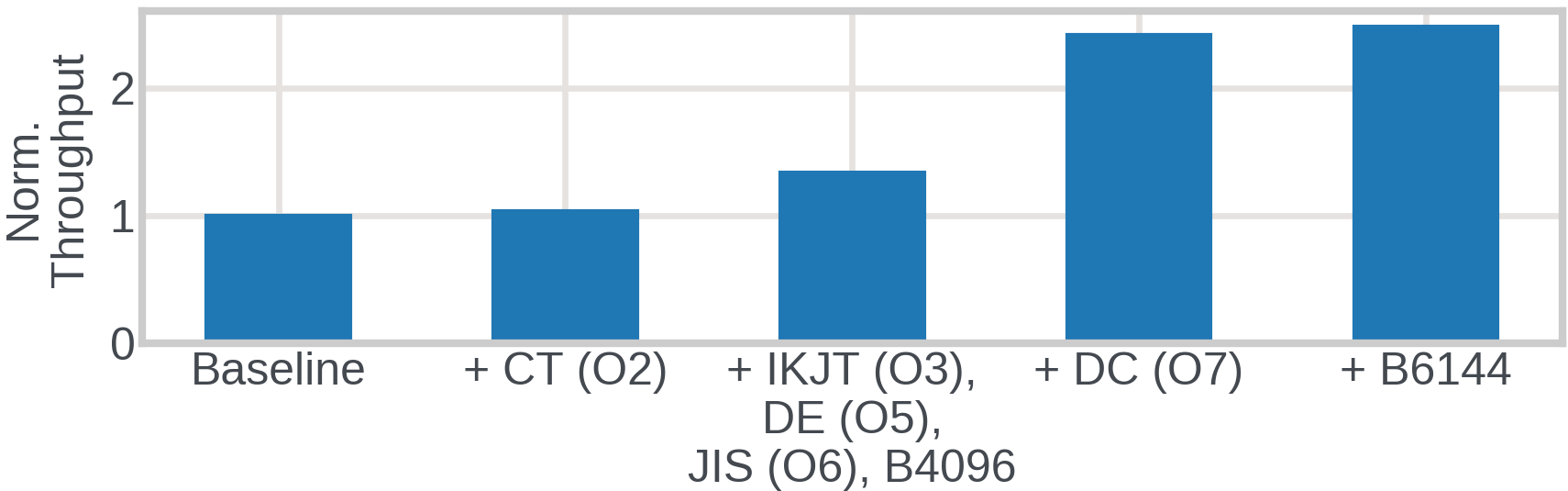}
  \vspace{-3mm}
  \caption{\small Ablation study showing normalized trainer throughput for $RM_1$. $O$=Table~\ref{tbl:optimizations} Optimization (O1 \& O4 not applicable). $CT$=Clustered Table, $DE$=Dedupe. EMB, $JIS$=Jagged Index Select, $DC$=Dedupe. Compute, $B$=Batch Size.}
  \label{fig:ablation}
\end{figure}

\noindent\textbf{Ablation Study.}
To understand how specific optimizations contribute to training throughput, we performed an ablation study using $RM_1$ as shown in Figure~\ref{fig:ablation}.
First, simply using clustered tables provides no training throughput benefit.
While clustering is necessary for \batchreuse, it is not sufficient alone since KJTs still contain duplicate feature values.
By using IKJTs (and jagged index select) to deduplicate EMB lookups and activations, we could increase batch size to 4096 and realized a $1.34\times$ gain to training throughput.
We then used multiple IKJT groups, which further allowed us to deduplicate the compute required by expensive transformers, which led to a $2.42\times$ increase in throughput.
Finally, this further allowed us to increase batch size to 6144, which resulted in a final $2.48\times$ increase in training throughput.

\noindent\textbf{Trainer Resource Utilization.}
\begin{table}[t]
\centering
\caption{Breakdown of trainer throughput (QPS) and efficiency for $RM_1$ with \batchreuse, enabling larger batches and complex models.}\label{tbl:hyperparams}
\footnotesize
\begin{tabular}{@{}l|rrrr@{}}
\toprule
Config. & \thead{Norm.\\QPS} & \thead{Max\\ Mem.\\ Util.} & \thead{Avg. \\Mem.\\ Util.} & \thead{Norm. Comp.\\ Efficiency \\ (flop/s/GPU)} \\ \midrule
Baseline         & 1.00          & 99.90  & 72.83 & 1.00          \\ \hline
RecD      & 1.89 & 27.76 & 22.20  & 1.73 \\ \hline
\thead[l]{RecD +\\ EMB D256} & 1.55 & 40.87 & 31.17 & 1.92 \\ \hline
\thead[l]{RecD +\\ B6144}       & 2.26 & 91.78 & 51.55 & 2.12 \\ \bottomrule
\end{tabular}%
\end{table}
Because \batchreuse reduces GPU resource requirements, we can also tune model hyper parameters in order to further improve model throughput and accuracy.
To illustrate this, Table~\ref{tbl:hyperparams} shows the throughput, memory utilization, and GPU compute efficiency for $RM_1$ as we enabled \batchreuse.
Using a baseline batch size of 2048 required the entirety of GPU memory. %
\batchreuse reduced the maximum and average memory utilization from $99.9\%$ to $27.76\%$ and $72.83\%$ to $22.2\%$, respectively.
This allowed us memory headroom to devote to EMBs or larger batches to improve model accuracy or training throughput, respectively.
For example, we were able to increase EMB dimensions from 128 to 256 or batch size from 2048 to 6144.
Furthermore, Table~\ref{tbl:hyperparams} shows how \batchreuse improves the utilization of GPU compute by increasing realized GPU FLOPS to $2.12\times$ the baseline.
GPU streaming multiprocessors can achieve higher utilization because they spend less time waiting for exposed A2A communication due to smaller IKJTs.

\revised{\noindent\textbf{Single-node Training.}
While \batchreuse greatly increased distributed training throughput, we also evaluated \batchreuse's benefit for single-node training.
To do so, we downsized $RM_1$ to fit within a single ZionEX training node and launched a training job with and without \batchreuse.
We observed a $2.18\times$ throughput increase in the single-node training setup by using \batchreuse.
\batchreuse still benefits single-node training because it targets GPU memory, network, and compute resources (Section~\ref{sec:training}).
While single-node training reduces the amount of exposed communication due to high-bandwidth NVLink interconnects, \batchreuse still improves compute and memory efficiency, leading to improved training throughput.
Since storage and readers are disaggregated, \batchreuse's benefits are the same for single-node training as shown in Figure~\ref{fig:model_eval}.
}

\noindent\textbf{Impacts to Accuracy.}
\batchreuse itself largely does not affect model accuracy.
Specifically, IKJTs encode the exact same logical data as KJTs and thus trainers can train on the exact same batches.
The only \batchreuse optimization that affects model accuracy is clustering tables by session ID.
In fact, clustering leads to significant improvements in accuracy.
This is because without clustering, duplicate examples from a session are distributed across batches.
The model observes the same sparse feature values for a user across multiple iterations, leading to multiple sparse updates.
This causes models to overfit for these features, negatively impacting generalization, especially for less popular (tail) values.
By grouping similar samples within the same batch, the model sees each user's data only once, reducing the chance of overfitting less popular sparse feature values.

\subsection{Why does \batchreuse improve reader throughput?}\label{sec:reader_eval}
\begin{figure}[t]
\centering
  \includegraphics[width=\linewidth]{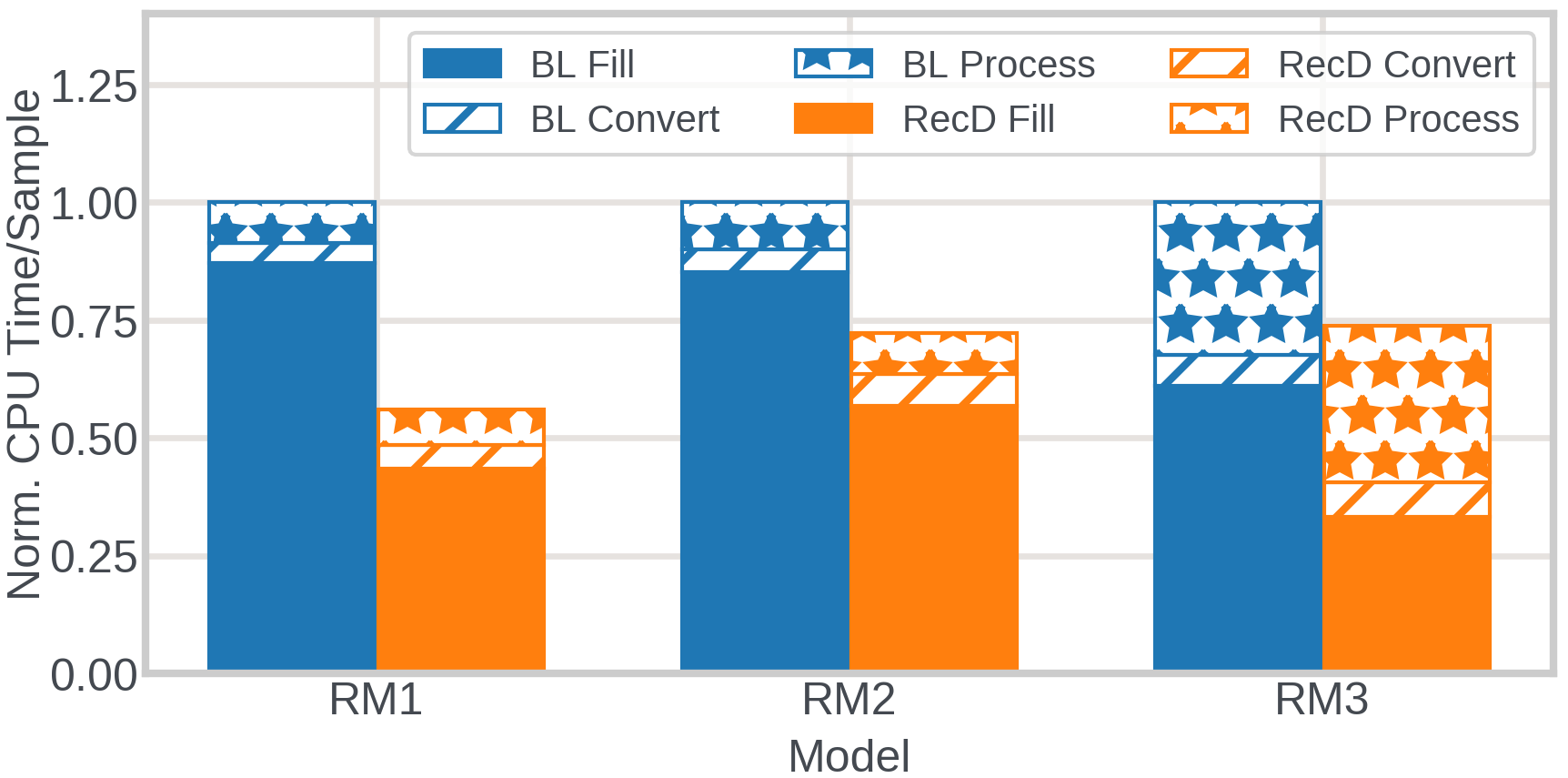}
    \vspace{-3mm}
  \caption{\small CPU time breakdown for a reader to preprocess a sample across $RM$s, normalized to each model's baseline CPU time.}
  \label{fig:reader_eval}
\end{figure}

\begin{table}[t]
\centering
\caption{Reader ingest \& egress bytes for a fixed \# of samples. }\label{tbl:reader}
\footnotesize
\begin{tabular}{@{}lll@{}}
\toprule
Experiment   & Read Bytes (GB) & Send Bytes (GB) \\ \midrule
Baseline     & 538             & 837             \\
with Cluster & 179             & 837             \\
with IKJT   & 179             & 713             \\ \bottomrule
\end{tabular}%
\end{table}
Figure~\ref{fig:model_eval} also showed how \batchreuse also improved the throughput of each reader, allowing us to provision fewer readers to feed trainers.
To understand why, Figure~\ref{fig:reader_eval} shows a breakdown of reader CPU compute time spent on \textit{filling}, \textit{converting} and \textit{processing} each sample, normalized to the baseline. %
For each $RM$, reader time is largely spent on fills: fetching data from \tectonic and decrypting, decompressing (zstd), and decoding bytes to form rows.
As shown in Table~\ref{tbl:reader}, by clustering tables, \batchreuse allows each reader to read significantly fewer bytes per sample.
Readers need to spend on both reading and extracting data, reducing the CPU time spent on fills by $50\%$, $33\%$, and $46\%$ for $RM_1$, $RM_2$, and $RM_3$, respectively.
Furthermore, \batchreuse can also reduce the CPU time required for processing, since preprocess operations take deduplicated IKJTs as input.
$RM_1$ and $RM_2$ required $13\%$ and $11\%$ less time for processing, while $RM_3$ was effectively the same ($3\%$ increase).

\batchreuse requires additional compute at readers to detect duplicate values (via hashing) during feature conversion. %
Fortunately, Figure~\ref{fig:reader_eval} shows that this overhead is largely negligible.
While the feature conversion time increased by $21\%$, $37\%$, and $11\%$ for $RM_1$, $RM_2$, and $RM_3$, respectively, feature conversion requires a small amount of overall compute.
Thus, the overall overhead of conversion is negligible ($1\%$) and is easily offset by fill and process benefits.

\subsection{Summary}
\begin{table}[]
\centering
\caption{\footnotesize \revised{Summary of impacts of each optimization from Table~\ref{tbl:optimizations} on end-to-end training pipeline performance for $RM_1$.}}\label{tbl:eval}
\footnotesize
\begin{tabular}{p{0.5cm}|p{6.8cm}}
\toprule
\textbf{Opt.} & \textbf{Effect on Performance}                                                                                                         \\ \midrule
\textbf{O1}   & \textbf{Storage}: Improves compression by 1.50x.                                                                                       \\ \hline
\textbf{O2}         & \textbf{Storage}: (with O1) Improves compression by 3.71x. \textbf{Reader}: Reduces CPU fill time by 50\% (improves reader throughput by 1.78x). \\ \hline
\textbf{O3}           & Enables O4-O6. \textbf{Reader}: Increases CPU convert time by 21\% (reduces reader throughput by 0.01x).                                \\ \hline
\textbf{O4}           & Enables O5-O6. \textbf{Reader}: Reduces CPU process time by 13\% (improves reader throughput by 0.01x).                                 \\ \hline
\textbf{O5}           & \textbf{Trainer}: (with O6 and B4096) Improves training throughput by 1.34x.                                                           \\ \hline
\textbf{O6}           & \textbf{Trainer}: (with O5 and B4096) Improves training throughput by 1.34x.                                                           \\ \hline
\textbf{O7}           & \textbf{Trainer}: (with O5 , O6, and B6144) Improves training throughput by 2.48x.                                                     \\ \bottomrule
\end{tabular}
\end{table}

\revised{
Table~\ref{tbl:eval} summarizes a breakdown of the impacts of each optimization presented in Table~\ref{tbl:optimizations} for the end-to-end training pipeline performance of $RM_1$, as reported in our evaluation results.
\batchreuse presents a suite of optimizations where not only does each optimization yield direct benefits at its respective pipeline system (e.g., storage, readers, or trainers), but it also enables further downstream optimizations.
O1 (sharding) and O2 (clustering) directly improve storage efficiency and reader throughput, but they also increase the opportunity for O3 (IKJTs) to deduplicate features.
While O3 introduces slight reader overheads, these are nullified by O4, and both O3 and O4 enable trainer-side optimizations.
These trainer-side optimizations (O5-O7) ultimately lead to a $2.48\times$ training throughput, as reported in Section~\ref{sec:trainer_eval}.
Collectively, \batchreuse optimizations benefit the end-to-end DLRM training pipeline, including storage, readers, and trainers.
}

\section{Discussion}\label{sec:discussion}
\noindent\textbf{Deciding Which Features to Deduplicate.}
ML engineers typically apply heuristics to decide which features to deduplicate.
IKJTs introduce no trainer overheads aside from an additional \texttt{index\_select} used to convert IKJTs to KJTs.
Thus, the benefit of deduplicating a feature $f$ must at least offset this overhead. %
While the specific ``worth it'' $DedupeFactor(f)$ threshold varies from model-to-model (due to model architecture differences illustrated in Section~\ref{sec:trainer_eval}), ML engineers will typically start by deduplicating features with $DedupeFactor(f) > 1.5$, and apply standard hyper parameter tuning techniques based on observed trainer throughput to finalize the deduplicated feature set.

\noindent\textbf{Boosting Dedupe Factors.}
$DedupeFactor(f)$ increases as a function of $S$, the average number of samples per session.
While Section~\ref{sec:characterization} showed that $S=16.5$ for the characterized industry-scale dataset, 
we are exploring methods to increase $S$ to yield benefits across the end-to-end training pipeline.
For example, the current data generation pipeline \textit{downsamples} (i.e., discards) training samples to keep datasets at a manageable size.
However, downsampling is applied on a per-sample basis.
By downsampling at a \textit{per-session} basis, we can further increase $S$, increasing $DedupeFactor$ without affecting model accuracy.

\noindent\textbf{Alternative Solutions and Generality.}
We considered alternative designs to exploit the session-centric characteristic of DLRM datasets.
One promising avenue was to explicitly deduplicate samples within the table schema itself.
Specifically, each user session requests inferences (one for each impression) in batches, and each batch uses the same features guaranteeing exact matches.
Instead of generating a table row for each impression, we considered generating one row (with one list for each feature) for each batch of impressions and encoding individual impressions as separate labels within a list.
This would directly deduplicate feature values by generating fewer rows within the dataset.

We decided against this approach for several reasons.
First, there would still be many duplicates as feature values are largely static even \textit{across} inference batches.
Secondly, we use a common dataset schema across DLRMs to ensure interoperability and developer velocity across models and datasets --- introducing a new schema would require significant engineering and adoption effort across multiple services.
\batchreuse is transparent to the training infrastructure because it does not require schema changes, and it enables even more deduplication across request batches.

Furthermore, \batchreuse optimizations are easily generalized to different environments, supporting myriad table schemas and model architectures.
Enabling \batchreuse reader and trainer optimizations only requires a \textit{feature converter} module to convert arbitrary table schemas into the IKJT encoding, and an \texttt{index\_select} call to convert IKJTs back to KJTs when necessary.
Because IKJTs directly build on standardized jagged tensors (ragged tensors in TensorFlow~\cite{website:ragged-tensor}), preprocessing functions and model architectures can operate on IKJTs as KJTs with minimal changes.

\noindent\textbf{Supporting Partial IKJTs.}
Supporting exact matches captures the vast majority of duplication in industry-scale datasets --- $81.6\%$ of an estimated $93.9\%$ maximum (Section~\ref{sec:characterization}).
Even so, IKJTs are also easily extended to support partial deduplication, capturing an additional $7.8\%$ of values, by leveraging the fact that partial matches are shifts.
Partial IKJTs remove the $offsets$ slice, and instead encode each row's $[offset, length]$ in the $inverse\_lookup$ slice.
In the example in Figure~\ref{fig:ikjt}, feature $b$ can by partially deduplicated via a partial IKJT consisting of $values=[3,4,5,6]$ and $inverse\_lookup=[[0,3], [1,3], [0,3]]$.
\section{Related Work}\label{sec:related}
\noindent\textbf{Duplication in DLRM Datasets.}
Gai et al. notes how DLRM datasets at Alibaba exhibits feature duplication~\cite{arxiv:gai_lsplm}.
To exploit this, the authors mention a ``common feature trick'' that routes samples from similar users to the same worker in a parameter server training setup.
The authors speed up training throughput by caching and reusing the parameter update for ``common'' features across each worker's samples.
Follow-up work by Ge et al. cites using the ``common feature trick'' during training~\cite{cikm18:ge_image-matters}.
Unfortunately, the authors provide scant details on how ``common'' features are generated, stored, or encoded.
They also do not elucidate how model architectures can exploit duplicate features, nor how the ``common feature trick'' can extend beyond parameter servers to synchronous training used in scale-out GPU training clusters~\cite{isca22:mudigere_zionex}.
We provide an in-depth characterization of feature value duplication in industrial DLRM datasets. %
\batchreuse deduplicates features across the end-to-end training pipeline by coalescing duplicate features in storage, compactly encoding them into IKJTs, and intelligently training on IKJTs.

\noindent\textbf{Deduplication in ML.}
Data deduplication has been studied in ML training outside of DLRMs.
Lee et al. studied how deduplication in a text corpus improved model accuracy for language tasks~\cite{acl22:lee_dedup}.
Allamanis studied how duplication in code datasets degraded model performance for ML models for source code~\cite{splash19:allamanis_code-dedup}.
To the best of our knowledge, our work is the first to study the systems implications of duplication in ML training datasets.

\noindent\textbf{Database Systems.}
Data deduplication is a well-studied area in databases.
IKJTs use a similar encoding mechanism to dictionary encoding commonly used in file formats such as Parquet~\cite{website:parquet}.
To coalesce duplicates within an IKJT, we rely on \textsc{cluster by} clauses supported by myriad database execution engines, such as Spark~\cite{nsdi12:zaharia_spark}.
\batchreuse applies these concepts to enable and encode deduplicated tensors for ML training jobs.

\noindent\textbf{Systems Optimizations for DLRM Training.}
Zhao et al. presented various optimizations to improve DSI efficiency for DLRM training at Meta~
\cite{isca22:zhao_dsi}.
RecShard~\cite{asplos22:sethi_recshard} and Adnan et al. \cite{vldb21:adnan_popular} leveraged skewed feature popularities to shard EMBs across GPUs, improving training throughput.
\revised{Similarly, Fleche~\cite{eurosys22:xie_fleche} is an embedding cache that caches EMBs on GPU HBM while relying on CPU DRAM for holding entire EMBs, targeting only single-GPU training.
To avoid cache write conflicts, Fleche recognizes that many sparse feature IDs are duplicated within a batch and performs only a single cache lookup for each unique ID, similar to \batchreuse's ability to deduplicate EMB lookups.}
TT-Rec~\cite{mlsys21:yin_ttrec} demonstrated compression techniques for EMBs.
\batchreuse provides orthogonal optimizations to improve storage, reading, and training performance by deduplicating features across the DLRM training pipeline.
\section{Conclusion}\label{sec:conclusion}
This paper presented \batchreuse, a suite of optimizations for industry-scale, end-to-end DLRM training pipelines.
We provide an in-depth characterization of how DLRM datasets exhibit inherent feature duplication.
\batchreuse coalesces duplicate features within a training batch, efficiently encodes them using IKJTs, and optimizes DLRM model architectures to train on deduplicated tensors.
As a result, \batchreuse improves training and preprocessing throughput and storage efficiency by up to $2.48\times$, $1.79\times$, and $3.71\times$, respectively.

\bibliography{paper}

\begin{thebibliography}{42}
\providecommand{\natexlab}[1]{#1}
\providecommand{\url}[1]{\texttt{#1}}
\expandafter\ifx\csname urlstyle\endcsname\relax
  \providecommand{\doi}[1]{doi: #1}\else
  \providecommand{\doi}{doi: \begingroup \urlstyle{rm}\Url}\fi

\bibitem[Acun et~al.(2021)Acun, Murphy, Wang, Nie, Wu, and
  Hazelwood]{acun:hpca21}
Acun, B., Murphy, M., Wang, X., Nie, J., Wu, C., and Hazelwood, K.
\newblock Understanding training efficiency of deep learning recommendation
  models at scale.
\newblock In \emph{2021 IEEE International Symposium on High-Performance
  Computer Architecture (HPCA)}, pp.\  802--814, 2021.

\bibitem[Adnan et~al.(2021)Adnan, Maboud, Mahajan, and
  Nair]{vldb21:adnan_popular}
Adnan, M., Maboud, Y.~E., Mahajan, D., and Nair, P.~J.
\newblock Accelerating recommendation system training by leveraging popular
  choices.
\newblock \emph{Proc. VLDB Endow.}, 15\penalty0 (1):\penalty0 127–140, sep
  2021.
\newblock ISSN 2150-8097.
\newblock \doi{10.14778/3485450.3485462}.
\newblock URL \url{https://doi.org/10.14778/3485450.3485462}.

\bibitem[Allamanis(2019)]{splash19:allamanis_code-dedup}
Allamanis, M.
\newblock The adverse effects of code duplication in machine learning models of
  code.
\newblock In \emph{Proceedings of the 2019 ACM SIGPLAN International Symposium
  on New Ideas, New Paradigms, and Reflections on Programming and Software},
  Onward! 2019, pp.\  143–153, New York, NY, USA, 2019. Association for
  Computing Machinery.
\newblock ISBN 9781450369954.
\newblock \doi{10.1145/3359591.3359735}.
\newblock URL \url{https://doi.org/10.1145/3359591.3359735}.

\bibitem[Anil et~al.(2022)Anil, Gadanho, Huang, Jacob, Li, Lin, Phillips, Pop,
  Regan, Shamir, Shivanna, and Yan]{orsum22:rohan_factoryfloor}
Anil, R., Gadanho, S., Huang, D., Jacob, N., Li, Z., Lin, D., Phillips, T.,
  Pop, C., Regan, K., Shamir, G.~I., Shivanna, R., and Yan, Q.
\newblock On the factory floor: Ml engineering for industrial-scale ads
  recommendation models, 2022.
\newblock URL \url{https://arxiv.org/abs/2209.05310}.

\bibitem[Apache(2022)]{website:parquet}
Apache.
\newblock Apache parquet file encodings, 2022.
\newblock URL
  \url{https://parquet.apache.org/docs/file-format/data-pages/encodings/}.

\bibitem[Ardalani et~al.(2022)Ardalani, Wu, Chen, Bhushanam, and
  Aziz]{arxiv:ardalani_scaling}
Ardalani, N., Wu, C.-J., Chen, Z., Bhushanam, B., and Aziz, A.
\newblock Understanding scaling laws for recommendation models, 2022.
\newblock URL \url{https://arxiv.org/abs/2208.08489}.

\bibitem[AWS(2022)]{website:aws_trainium}
AWS.
\newblock Aws trainium, 2022.
\newblock URL \url{https://aws.amazon.com/machine-learning/trainium/}.

\bibitem[Chen et~al.(2019)Chen, Zhao, Li, Huang, and
  Ou]{dlpkdd19:chen_behavior-sequence-transformer}
Chen, Q., Zhao, H., Li, W., Huang, P., and Ou, W.
\newblock Behavior sequence transformer for e-commerce recommendation in
  alibaba.
\newblock In \emph{Proceedings of the 1st International Workshop on Deep
  Learning Practice for High-Dimensional Sparse Data}, DLP-KDD '19, New York,
  NY, USA, 2019. Association for Computing Machinery.
\newblock ISBN 9781450367837.
\newblock \doi{10.1145/3326937.3341261}.
\newblock URL \url{https://doi.org/10.1145/3326937.3341261}.

\bibitem[de~Souza Pereira~Moreira et~al.(2021)de~Souza Pereira~Moreira, Rabhi,
  Lee, Ak, and Oldridge]{recsys21:desouza_transformers4rec}
de~Souza Pereira~Moreira, G., Rabhi, S., Lee, J.~M., Ak, R., and Oldridge, E.
\newblock Transformers4rec: Bridging the gap between nlp and sequential /
  session-based recommendation.
\newblock In \emph{Proceedings of the 15th ACM Conference on Recommender
  Systems}, RecSys '21, pp.\  143–153, New York, NY, USA, 2021. Association
  for Computing Machinery.
\newblock ISBN 9781450384582.
\newblock \doi{10.1145/3460231.3474255}.
\newblock URL \url{https://doi.org/10.1145/3460231.3474255}.

\bibitem[Fang et~al.(2020)Fang, Zhang, Shu, and
  Guo]{transinfsyst21:fang_sequential}
Fang, H., Zhang, D., Shu, Y., and Guo, G.
\newblock Deep learning for sequential recommendation: Algorithms, influential
  factors, and evaluations.
\newblock \emph{ACM Trans. Inf. Syst.}, 39\penalty0 (1), nov 2020.
\newblock ISSN 1046-8188.
\newblock \doi{10.1145/3426723}.
\newblock URL \url{https://doi.org/10.1145/3426723}.

\bibitem[Gai et~al.(2017)Gai, Zhu, Li, Liu, and Wang]{arxiv:gai_lsplm}
Gai, K., Zhu, X., Li, H., Liu, K., and Wang, Z.
\newblock Learning piece-wise linear models from large scale data for ad click
  prediction, 2017.
\newblock URL \url{https://arxiv.org/abs/1704.05194}.

\bibitem[Ge et~al.(2018)Ge, Zhao, Zhou, Chen, Liu, Yi, Hu, Liu, Sun, Liu, Yi,
  Huang, Zhang, Zhu, Zhang, and Gai]{cikm18:ge_image-matters}
Ge, T., Zhao, L., Zhou, G., Chen, K., Liu, S., Yi, H., Hu, Z., Liu, B., Sun,
  P., Liu, H., Yi, P., Huang, S., Zhang, Z., Zhu, X., Zhang, Y., and Gai, K.
\newblock Image matters: Visually modeling user behaviors using advanced model
  server.
\newblock In \emph{Proceedings of the 27th ACM International Conference on
  Information and Knowledge Management}, CIKM '18, pp.\  2087–2095, New York,
  NY, USA, 2018. Association for Computing Machinery.
\newblock ISBN 9781450360142.
\newblock \doi{10.1145/3269206.3272007}.
\newblock URL \url{https://doi.org/10.1145/3269206.3272007}.

\bibitem[Hazelwood et~al.(2018)Hazelwood, Bird, Brooks, Chintala, Diril,
  Dzhulgakov, Fawzy, Jia, Jia, Kalro, Law, Lee, Lu, Noordhuis, Smelyanskiy,
  Xiong, and Wang]{hpca18:hazelwood_applied-ml}
Hazelwood, K., Bird, S., Brooks, D., Chintala, S., Diril, U., Dzhulgakov, D.,
  Fawzy, M., Jia, B., Jia, Y., Kalro, A., Law, J., Lee, K., Lu, J., Noordhuis,
  P., Smelyanskiy, M., Xiong, L., and Wang, X.
\newblock Applied machine learning at facebook: A datacenter infrastructure
  perspective.
\newblock In \emph{2018 IEEE International Symposium on High Performance
  Computer Architecture (HPCA)}, pp.\  620--629, 2018.
\newblock \doi{10.1109/HPCA.2018.00059}.

\bibitem[Karpathiotakis et~al.(2019)Karpathiotakis, Wernli, and
  Stojanovic]{website:scribe}
Karpathiotakis, M., Wernli, D., and Stojanovic, M.
\newblock Scribe: Transporting petabytes per hour via a distributed, buffered
  queueing system, Oct 2019.
\newblock URL
  \url{https://engineering.fb.com/2019/10/07/data-infrastructure/scribe/}.

\bibitem[Kaufman et~al.(2012)Kaufman, Rosset, Perlich, and
  Stitelman]{tkdd12:kaufman_leakage}
Kaufman, S., Rosset, S., Perlich, C., and Stitelman, O.
\newblock Leakage in data mining: Formulation, detection, and avoidance.
\newblock \emph{ACM Trans. Knowl. Discov. Data}, 6\penalty0 (4), dec 2012.
\newblock ISSN 1556-4681.
\newblock \doi{10.1145/2382577.2382579}.
\newblock URL \url{https://doi.org/10.1145/2382577.2382579}.

\bibitem[Lardinois(2022)]{website:tpuv4-cloud}
Lardinois, F.
\newblock Google launches a 9 exaflop cluster of cloud {TPU} v4 pods into
  public preview.
\newblock
  \url{https://techcrunch.com/2022/05/11/google-launches-a-9-exaflop-cluster-of-cloud-tpu-v4-pods-into-public-preview/},
  May 2022.

\bibitem[Lee et~al.(2022)Lee, Ippolito, Nystrom, Zhang, Eck, Callison-Burch,
  and Carlini]{acl22:lee_dedup}
Lee, K., Ippolito, D., Nystrom, A., Zhang, C., Eck, D., Callison-Burch, C., and
  Carlini, N.
\newblock Deduplicating training data makes language models better.
\newblock In \emph{Proceedings of the 60th Annual Meeting of the Association
  for Computational Linguistics}. Association for Computational Linguistics,
  2022.

\bibitem[Li et~al.(2019)Li, Liu, Wu, Xu, Zhao, Huang, Kang, Chen, Li, and
  Lee]{cikm19:li_tmall}
Li, C., Liu, Z., Wu, M., Xu, Y., Zhao, H., Huang, P., Kang, G., Chen, Q., Li,
  W., and Lee, D.~L.
\newblock Multi-interest network with dynamic routing for recommendation at
  tmall.
\newblock In \emph{Proceedings of the 28th ACM International Conference on
  Information and Knowledge Management}, CIKM '19, pp.\  2615–2623, New York,
  NY, USA, 2019. Association for Computing Machinery.
\newblock ISBN 9781450369763.
\newblock \doi{10.1145/3357384.3357814}.
\newblock URL \url{https://doi.org/10.1145/3357384.3357814}.

\bibitem[Li et~al.(2020)Li, Qin, Wang, Chen, and
  Metzler]{www20:li_search-ranking}
Li, R., Qin, Z., Wang, X., Chen, S.~J., and Metzler, D.
\newblock \emph{Stabilizing Neural Search Ranking Models}, pp.\  2725–2732.
\newblock Association for Computing Machinery, New York, NY, USA, 2020.
\newblock ISBN 9781450370233.
\newblock URL \url{https://doi.org/10.1145/3366423.3380030}.

\bibitem[Meta(2019)]{website:instgram-ranking}
Meta.
\newblock Powered by ai: Instagram's explore recommender system, 2019.
\newblock URL
  \url{https://ai.facebook.com/blog/powered-by-ai-instagrams-explore-recommender-system/}.

\bibitem[Meta(2022)]{meta:rsc}
Meta.
\newblock Introducing the ai research supercluster, 2022.
\newblock URL \url{https://ai.facebook.com/blog/ai-rsc/}.

\bibitem[Mudigere et~al.(2022)Mudigere, Hao, Huang, Jia, Tulloch, Sridharan,
  Liu, Ozdal, Nie, Park, Luo, Yang, Gao, Ivchenko, Basant, Hu, Yang, Ardestani,
  Wang, Komuravelli, Chu, Yilmaz, Li, Qian, Feng, Ma, Yang, Wen, Li, Yang, Sun,
  Zhao, Melts, Dhulipala, Kishore, Graf, Eisenman, Matam, Gangidi, Chen,
  Krishnan, Nayak, Nair, Muthiah, khorashadi, Bhattacharya, Lapukhov, Naumov,
  Mathews, Qiao, Smelyanskiy, Jia, and Rao]{isca22:mudigere_zionex}
Mudigere, D., Hao, Y., Huang, J., Jia, Z., Tulloch, A., Sridharan, S., Liu, X.,
  Ozdal, M., Nie, J., Park, J., Luo, L., Yang, J.~A., Gao, L., Ivchenko, D.,
  Basant, A., Hu, Y., Yang, J., Ardestani, E.~K., Wang, X., Komuravelli, R.,
  Chu, C.-H., Yilmaz, S., Li, H., Qian, J., Feng, Z., Ma, Y., Yang, J., Wen,
  E., Li, H., Yang, L., Sun, C., Zhao, W., Melts, D., Dhulipala, K., Kishore,
  K., Graf, T., Eisenman, A., Matam, K.~K., Gangidi, A., Chen, G.~J., Krishnan,
  M., Nayak, A., Nair, K., Muthiah, B., khorashadi, M., Bhattacharya, P.,
  Lapukhov, P., Naumov, M., Mathews, A., Qiao, L., Smelyanskiy, M., Jia, B.,
  and Rao, V.
\newblock Software-hardware co-design for fast and scalable training of deep
  learning recommendation models.
\newblock In \emph{Proceedings of the 49th Annual International Symposium on
  Computer Architecture}, ISCA '22, pp.\  993–1011, New York, NY, USA, 2022.
  Association for Computing Machinery.
\newblock ISBN 9781450386104.
\newblock \doi{10.1145/3470496.3533727}.
\newblock URL \url{https://doi.org/10.1145/3470496.3533727}.

\bibitem[Naumov et~al.(2019)Naumov, Mudigere, Shi, Huang, Sundaraman, Park,
  Wang, Gupta, Wu, Azzolini, Dzhulgakov, Mallevich, Cherniavskii, Lu,
  Krishnamoorthi, Yu, Kondratenko, Pereira, Chen, Chen, Rao, Jia, Xiong, and
  Smelyanskiy]{arxiv:naumov_dlrm}
Naumov, M., Mudigere, D., Shi, H.-J.~M., Huang, J., Sundaraman, N., Park, J.,
  Wang, X., Gupta, U., Wu, C.-J., Azzolini, A.~G., Dzhulgakov, D., Mallevich,
  A., Cherniavskii, I., Lu, Y., Krishnamoorthi, R., Yu, A., Kondratenko, V.,
  Pereira, S., Chen, X., Chen, W., Rao, V., Jia, B., Xiong, L., and
  Smelyanskiy, M.
\newblock Deep learning recommendation model for personalization and
  recommendation systems, 2019.
\newblock URL \url{https://arxiv.org/abs/1906.00091}.

\bibitem[Naumov et~al.(2020)Naumov, Kim, Mudigere, Sridharan, Wang, Zhao,
  Yilmaz, Kim, Yuen, Ozdal, Nair, Gao, Su, Yang, and
  Smelyanskiy]{arxiv:naumov_fb-dlt}
Naumov, M., Kim, J., Mudigere, D., Sridharan, S., Wang, X., Zhao, W., Yilmaz,
  S., Kim, C., Yuen, H., Ozdal, M., Nair, K., Gao, I., Su, B.-Y., Yang, J., and
  Smelyanskiy, M.
\newblock Deep learning training in facebook data centers: Design of scale-up
  and scale-out systems, 2020.
\newblock URL \url{https://arxiv.org/abs/2003.09518}.

\bibitem[NVIDIA(2022)]{nvidia:nccl}
NVIDIA.
\newblock Nvidia nccl, 2022.
\newblock URL \url{https://developer.nvidia.com/nccl}.

\bibitem[ORC(2022)]{website:apache-orc}
ORC, A.
\newblock Apache orc: High-performance columnar storage for hadoop, 2022.
\newblock URL \url{https://orc.apache.org/}.

\bibitem[Pan et~al.(2021)Pan, Stavrinos, Zhang, Sikaria, Zakharov, Sharma, P,
  Shuey, Wareing, Gangapuram, Cao, Preseau, Singh, Patiejunas, Tipton,
  Katz-Bassett, and Lloyd]{fast21:pan_tectonic}
Pan, S., Stavrinos, T., Zhang, Y., Sikaria, A., Zakharov, P., Sharma, A., P,
  S.~S., Shuey, M., Wareing, R., Gangapuram, M., Cao, G., Preseau, C., Singh,
  P., Patiejunas, K., Tipton, J., Katz-Bassett, E., and Lloyd, W.
\newblock Facebook{\textquoteright}s tectonic filesystem: Efficiency from
  exascale.
\newblock In \emph{19th {USENIX} Conference on File and Storage Technologies
  ({FAST} 21)}, pp.\  217--231. {USENIX} Association, February 2021.
\newblock ISBN 978-1-939133-20-5.
\newblock URL \url{https://www.usenix.org/conference/fast21/presentation/pan}.

\bibitem[Patterson et~al.(2022)Patterson, Gonzalez, Hölzle, Le, Liang,
  Munguia, Rothchild, So, Texier, and Dean]{computer22:patterson_carbon}
Patterson, D., Gonzalez, J., Hölzle, U., Le, Q., Liang, C., Munguia, L.-M.,
  Rothchild, D., So, D.~R., Texier, M., and Dean, J.
\newblock The carbon footprint of machine learning training will plateau, then
  shrink.
\newblock \emph{Computer}, 55\penalty0 (7):\penalty0 18--28, 2022.
\newblock \doi{10.1109/MC.2022.3148714}.

\bibitem[Pi et~al.(2019)Pi, Bian, Zhou, Zhu, and Gai]{kdd19:pi_uic}
Pi, Q., Bian, W., Zhou, G., Zhu, X., and Gai, K.
\newblock Practice on long sequential user behavior modeling for click-through
  rate prediction.
\newblock In \emph{Proceedings of the 25th ACM SIGKDD International Conference
  on Knowledge Discovery \& Data Mining}, KDD '19, pp.\  2671–2679, New York,
  NY, USA, 2019. Association for Computing Machinery.
\newblock ISBN 9781450362016.
\newblock \doi{10.1145/3292500.3330666}.
\newblock URL \url{https://doi.org/10.1145/3292500.3330666}.

\bibitem[PyTorch(2022)]{website:torchrec-sparse}
PyTorch.
\newblock Torchrec.sparse, 2022.
\newblock URL \url{https://pytorch.org/torchrec/torchrec.sparse.html}.

\bibitem[PyTorch(2023)]{website:pytorch-dataloader}
PyTorch.
\newblock torch.utils.data, 2023.
\newblock URL \url{https://pytorch.org/docs/stable/data.html}.

\bibitem[Sethi et~al.(2022)Sethi, Acun, Agarwal, Kozyrakis, Trippel, and
  Wu]{asplos22:sethi_recshard}
Sethi, G., Acun, B., Agarwal, N., Kozyrakis, C., Trippel, C., and Wu, C.-J.
\newblock Recshard: Statistical feature-based memory optimization for
  industry-scale neural recommendation.
\newblock In \emph{Proceedings of the 27th ACM International Conference on
  Architectural Support for Programming Languages and Operating Systems},
  ASPLOS 2022, pp.\  344–358, New York, NY, USA, 2022. Association for
  Computing Machinery.
\newblock ISBN 9781450392051.
\newblock \doi{10.1145/3503222.3507777}.
\newblock URL \url{https://doi.org/10.1145/3503222.3507777}.

\bibitem[TensorFlow(2022)]{website:ragged-tensor}
TensorFlow.
\newblock Ragged tensors, 2022.
\newblock URL \url{https://www.tensorflow.org/guide/ragged_tensor}.

\bibitem[Thusoo et~al.(2009)Thusoo, Sarma, Jain, Shao, Chakka, Anthony, Liu,
  Wyckoff, and Murthy]{vldb09:thusoo_hive}
Thusoo, A., Sarma, J.~S., Jain, N., Shao, Z., Chakka, P., Anthony, S., Liu, H.,
  Wyckoff, P., and Murthy, R.
\newblock Hive: A warehousing solution over a map-reduce framework.
\newblock \emph{Proc. VLDB Endow.}, 2\penalty0 (2):\penalty0 1626–1629, aug
  2009.
\newblock ISSN 2150-8097.
\newblock \doi{10.14778/1687553.1687609}.
\newblock URL \url{https://doi.org/10.14778/1687553.1687609}.

\bibitem[Vaswani et~al.(2017)Vaswani, Shazeer, Parmar, Uszkoreit, Jones, Gomez,
  Kaiser, and Polosukhin]{neurips17:vaswani_attention}
Vaswani, A., Shazeer, N., Parmar, N., Uszkoreit, J., Jones, L., Gomez, A.~N.,
  Kaiser, L.~u., and Polosukhin, I.
\newblock Attention is all you need.
\newblock In Guyon, I., Luxburg, U.~V., Bengio, S., Wallach, H., Fergus, R.,
  Vishwanathan, S., and Garnett, R. (eds.), \emph{Advances in Neural
  Information Processing Systems}, volume~30. Curran Associates, Inc., 2017.
\newblock URL
  \url{https://proceedings.neurips.cc/paper/2017/file/3f5ee243547dee91fbd053c1c4a845aa-Paper.pdf}.

\bibitem[Wang et~al.(2021)Wang, Cao, Wang, Sheng, Orgun, and
  Lian]{survey21:wang_sbrs}
Wang, S., Cao, L., Wang, Y., Sheng, Q.~Z., Orgun, M.~A., and Lian, D.
\newblock A survey on session-based recommender systems.
\newblock \emph{ACM Comput. Surv.}, 54\penalty0 (7), jul 2021.
\newblock ISSN 0360-0300.
\newblock \doi{10.1145/3465401}.
\newblock URL \url{https://doi.org/10.1145/3465401}.

\bibitem[Xie et~al.(2022)Xie, Lu, Lin, Wang, Gao, Ren, and
  Shu]{eurosys22:xie_fleche}
Xie, M., Lu, Y., Lin, J., Wang, Q., Gao, J., Ren, K., and Shu, J.
\newblock Fleche: An efficient gpu embedding cache for personalized
  recommendations.
\newblock In \emph{Proceedings of the Seventeenth European Conference on
  Computer Systems}, EuroSys '22, pp.\  402–416, New York, NY, USA, 2022.
  Association for Computing Machinery.
\newblock ISBN 9781450391627.
\newblock \doi{10.1145/3492321.3519554}.
\newblock URL \url{https://doi.org/10.1145/3492321.3519554}.

\bibitem[Yin et~al.(2021)Yin, Acun, Wu, and Liu]{mlsys21:yin_ttrec}
Yin, C., Acun, B., Wu, C.-J., and Liu, X.
\newblock Tt-rec: Tensor train compression for deep learning recommendation
  models.
\newblock \emph{Proceedings of Machine Learning and Systems}, 3:\penalty0
  448--462, 2021.

\bibitem[Zaharia et~al.(2012)Zaharia, Chowdhury, Das, Dave, Ma, McCauly,
  Franklin, Shenker, and Stoica]{nsdi12:zaharia_spark}
Zaharia, M., Chowdhury, M., Das, T., Dave, A., Ma, J., McCauly, M., Franklin,
  M.~J., Shenker, S., and Stoica, I.
\newblock Resilient distributed datasets: A fault-tolerant abstraction for
  in-memory cluster computing.
\newblock In \emph{9th {USENIX} Symposium on Networked Systems Design and
  Implementation ({NSDI} 12)}, pp.\  15--28, San Jose, CA, April 2012. {USENIX}
  Association.
\newblock ISBN 978-931971-92-8.
\newblock URL
  \url{https://www.usenix.org/conference/nsdi12/technical-sessions/presentation/zaharia}.

\bibitem[Zhao et~al.(2022)Zhao, Agarwal, Basant, Gedik, Pan, Ozdal,
  Komuravelli, Pan, Bao, Lu, Narayanan, Langman, Wilfong, Rastogi, Wu,
  Kozyrakis, and Pol]{isca22:zhao_dsi}
Zhao, M., Agarwal, N., Basant, A., Gedik, B., Pan, S., Ozdal, M., Komuravelli,
  R., Pan, J., Bao, T., Lu, H., Narayanan, S., Langman, J., Wilfong, K.,
  Rastogi, H., Wu, C.-J., Kozyrakis, C., and Pol, P.
\newblock Understanding data storage and ingestion for large-scale deep
  recommendation model training: Industrial product.
\newblock In \emph{Proceedings of the 49th Annual International Symposium on
  Computer Architecture}, ISCA '22, pp.\  1042–1057, New York, NY, USA, 2022.
  Association for Computing Machinery.
\newblock ISBN 9781450386104.
\newblock \doi{10.1145/3470496.3533044}.
\newblock URL \url{https://doi.org/10.1145/3470496.3533044}.

\bibitem[Zhao et~al.(2019)Zhao, Hong, Wei, Chen, Nath, Andrews, Kumthekar,
  Sathiamoorthy, Yi, and Chi]{recsys19:zhao_youtube}
Zhao, Z., Hong, L., Wei, L., Chen, J., Nath, A., Andrews, S., Kumthekar, A.,
  Sathiamoorthy, M., Yi, X., and Chi, E.
\newblock Recommending what video to watch next: A multitask ranking system.
\newblock In \emph{Proceedings of the 13th ACM Conference on Recommender
  Systems}, RecSys '19, pp.\  43–51, New York, NY, USA, 2019. Association for
  Computing Machinery.
\newblock ISBN 9781450362436.
\newblock \doi{10.1145/3298689.3346997}.
\newblock URL \url{https://doi.org/10.1145/3298689.3346997}.

\bibitem[Zstandard(2022)]{website:zstd}
Zstandard.
\newblock Zstandard, 2022.
\newblock URL \url{www.zstd.net}.

\end{thebibliography}
\bibliographystyle{mlsys2023}

\end{document}